\documentclass[manuscript,screen]  
{acmart}  
\usepackage{multirow}
\usepackage{pifont} 
\usepackage{wrapfig}
\usepackage{array}
\usepackage{xcolor}
\usepackage{longtable}
\usepackage{geometry}
\usepackage{MnSymbol,bbding,pifont}
\renewcommand\footnotetextcopyrightpermission[1]{} 
\AtBeginDocument{%
  }

\setcopyright{acmlicensed}
\copyrightyear{2018}
\acmYear{2018}
\acmDOI{XXXXXXX.XXXXXXX}
\acmConference[Conference acronym 'XX]{Make sure to enter the correct
  conference title from your rights confirmation email}{June 03--05,
  2018}{Woodstock, NY}
\acmISBN{978-1-4503-XXXX-X/2018/06}

\acmSubmissionID{CSUR-2025-0899}

\begin{document}

\title{A Survey: Spatiotemporal Consistency in Video Generation}

\author{Zhiyu Yin}
\email{ZhiyuYin88@gmail.com}
\orcid{0009-0003-5280-1280}
\affiliation{%
  \institution{School of Computer Science and Technology, Harbin Institute of Technology, Shenzhen}
  \city{Shenzhen}
  \country{China}
}
\affiliation{%
  \institution{School of Computer Science and Engineering, Central South University}
  \city{Changsha}
  \country{China}
}

\author{Kehai Chen}
\affiliation{%
  \institution{School of Computer Science and Technology, Harbin Institute of Technology, Shenzhen}
  \city{Shenzhen}
  \country{China}
}
\affiliation{%
  \institution{Peng Cheng Laboratory}
  \city{Shenzhen}
  \country{China}
}
\email{chenkehai@hit.edu.cn}

\author{Xuefeng Bai}
\affiliation{%
  \institution{School of Computer Science and Technology, Harbin Institute of Technology, Shenzhen}
  \city{Shenzhen}
  \country{China}
}
\email{baixuefeng@hit.edu.cn}

\author{Ruili Jiang}
\affiliation{%
  \institution{School of Computer Science and Technology, Harbin Institute of Technology, Shenzhen}
  \city{Shenzhen}
  \country{China}
}
\email{ruilijiang@outlook.com}

\author{Juntao Li}
\affiliation{%
  \institution{School of Computer Science and Technology, Soochow University}
  \city{Suzhou}
  \country{China}
}
\email{ljt@suda.edu.cn}

\author{Hong-Dong Li}
\affiliation{%
  \institution{School of Computer Science and Engineering, Central South University}
  \city{Changsha}
  \country{China}
}
\email{hongdong@csu.edu.cn}

\author{Jin Liu}
\affiliation{%
  \institution{School of Computer Science and Engineering, Central South University}
  \city{Changsha}
  \country{China}
}
\email{liujin06@csu.edu.cn}

\author{Yang Xiang}
\affiliation{%
  \institution{Peng Cheng Laboratory}
  \city{Shenzhen}
  \country{China}
}
\email{xiangy@pcl.ac.cn}

\author{Jun Yu}
\affiliation{%
  \institution{School of Computer Science and Technology, Harbin Institute of Technology, Shenzhen}
  \city{Shenzhen}
  \country{China}
}
\email{yujun@hit.edu.cn}

\author{Min Zhang}
\affiliation{%
  \institution{School of Computer Science and Technology, Harbin Institute of Technology, Shenzhen}
  \city{Shenzhen}
  \country{China}
}
\affiliation{%
  \institution{Peng Cheng Laboratory}
  \city{Shenzhen}
  \country{China}
}
\email{zhangmin2021@hit.edu.cn}

\renewcommand{\shortauthors}{Yin et al.}

\begin{abstract}
  Video generation aims to produce temporally coherent sequences of visual frames, representing a pivotal advancement in Artificial Intelligence Generated Content (AIGC).
  Compared to static image generation, video generation poses unique challenges: it demands not only high-quality individual frames but also strong temporal coherence to ensure consistency throughout the spatiotemporal sequence. Although research addressing spatiotemporal consistency in video generation has increased in recent years, systematic reviews focusing on this core issue remain relatively scarce. To fill this gap, this paper views the video generation task as a sequential sampling process from a high-dimensional spatiotemporal distribution, and further discusses spatiotemporal consistency. We provide a systematic review of the latest advancements in the field. The content spans multiple dimensions including generation models, feature representations, generation frameworks, post-processing techniques, training strategies, benchmarks and evaluation metrics, with a particular focus on the mechanisms and effectiveness of various methods in maintaining spatiotemporal consistency. Finally, this paper explores future research directions and potential challenges in this field, aiming to provide valuable insights for advancing video generation technology. The project link is \href{https://github.com/Yin-Z-Y/A-Survey-Spatiotemporal-Consistency-in-Video-Generation}{https://github.com/Yin-Z-Y/A-Survey-Spatiotemporal-Consistency-in-Video-Generation}

\end{abstract}




\maketitle

\section{Introduction}
\label{Sec:Introduction}
         Artificial Intelligence Generated Content (AIGC)~\cite{yang2023diffusion,li2024survey,liu2024sora} has become a mainstream application of AI, leveraging algorithms and models to create new content that meets user needs, greatly impacting people's production and daily life. Among them, video, as an important form of visual content, rapidly displays a series of images or frames in succession, creating the perception of motion. It offers a rich and dynamic medium for expressing and transmitting information. In recent years, there have been transformative advancements in video generation fields~\cite{yang2023diffusion,li2024survey,liu2024sora,melnik2024video,xiong2024autoregressive}. Major AI institutions and companies have invested substantial resources into developing video generation products that meet basic level, such as Runway's Gen series~\cite{gen-1,gen-2,gen-3,esser2023structure}, Google DeepMind's Veo~\cite{veo} and OpenAI's Sora~\cite{sora}.\par


         In current research on video generation, scholars generally conceptualize video as high-dimensional spatiotemporal data~\cite{ho2022imagen,hong2022cogvideo}. Its generation process can be understood as sampling from an implicit high-dimensional spatiotemporal probability distribution under constraints such as text and reference images, thereby producing dynamic sequences that meet semantic and visual requirements. Specifically, video can be decomposed into a sequence of spatiotemporal units, where each unit may represent a frame, a local frame block, or a feature block in the latent space. The entire video is then constructed by sequentially connecting these units. The key challenge in generation lies in maintaining spatiotemporal consistency among these units~\cite{tran2015learning,zhou2019spatio,yang2023diffusion,bar2024lumiere,li2024survey,liu2024sora,xiong2024autoregressive}.
         
\begin{figure*}[t!]
            \centering
            \includegraphics[width=\textwidth]{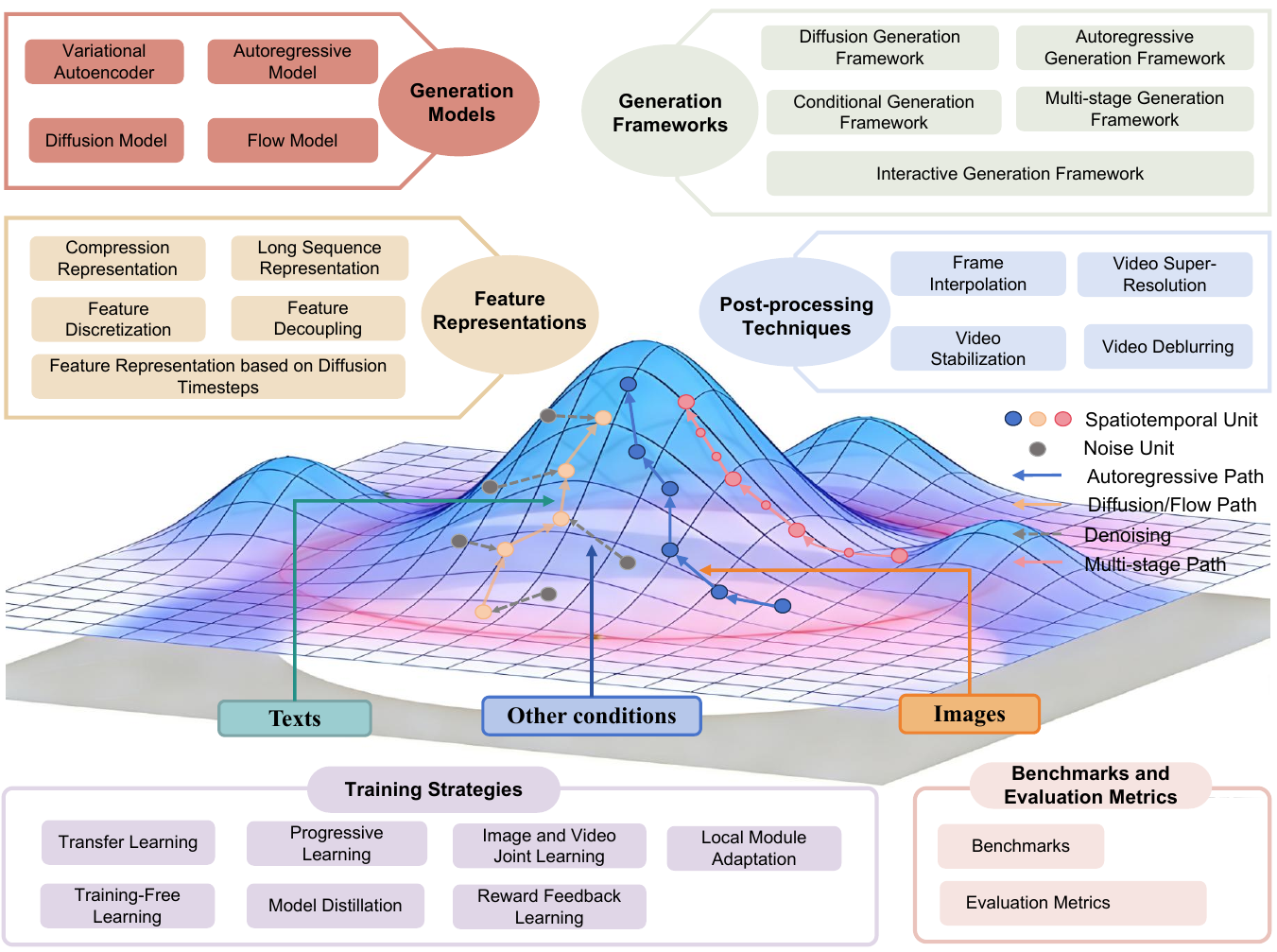}
            
            \caption{Overview Structure Diagram.}
            \label{Fig:Overview Structure Diagram}
            \vspace{-1.2cm}
        \end{figure*}

          However, "spatiotemporal consistency" as an abstract concept is difficult to directly elaborate upon. Our work proposes reformulating the problem from the perspective of sequence sampling in high-dimensional spatiotemporal distributions (See Figure \ref{Fig:Overview Structure Diagram} ). We assume the existence of a spatiotemporal feature distribution $p(V)$ that captures the intrinsic structure of video data, where each spatiotemporal unit $x_t$ can be regarded as a sample point drawn from this distribution, and the complete video $V = [x_1, x_2, ..., x_T]$ constitutes the temporal sequence formed by these sample points. Within this framework, \textbf{Spatial Consistency} is defined as the compatibility between any two sample points in the distribution regarding semantic and visual attributes, including stability in subject identity, scene layout, lighting style, and so forth. \textbf{Temporal Consistency} further requires smooth evolutionary transitions between adjacent samples in the sequence. Even if each unit independently follows the spatial consistency, failure to adhere to reasonable temporal dynamics during connection will still cause issues like image jumps, flickering, or unnatural motion. Therefore, temporal consistency is fundamentally a transition probability modeling problem for sequence generation—specifically, learning the conditional distribution $p(x_t|x_{<t},C)$, where $C$ represents control information. In Table \ref{tab:Classification of Spatiotemporal Consistency}, we categorize and summarize some common spatiotemporal inconsistencies encountered in the field of video generation. The appendix \ref{Fig:Cases of Spatiotemporal Inconsistency} presents several generated examples to better illustrate these issues.


\begin{table}[]
\centering
\scalebox{0.65}{
\begin{tabular}{c|c|l|c}
\hline
\textbf{Consistency Dimension}                 & \textbf{Categories}           & \multicolumn{1}{c|}{\textbf{Description}}                                                                                                                              & \textbf{Common Issues}                                                               \\ \hline
\multirow{5}{*}{\textbf{Spatial Consistency}}  & Subject Identity Consistency  & Primary subjects maintain stability in appearance, form, and identity.                                                                                                 & \begin{tabular}[c]{@{}c@{}}Subject swapping\\ Color change\end{tabular}              \\ \cline{2-4} 
                                               & Scene Layout Consistency      & \begin{tabular}[c]{@{}l@{}}Background, relative positions, and spatial structures remain consistent\\  across different units.\end{tabular}                            & Background switching                                                                 \\ \cline{2-4} 
                                               & Lighting \& Style Consistency & \begin{tabular}[c]{@{}l@{}}Maintain consistency in overall lighting direction, intensity, color tone, \\ and artistic style.\end{tabular}                              & \begin{tabular}[c]{@{}c@{}}Lighting flicker\\ Abrupt changes in style\end{tabular}   \\ \cline{2-4} 
                                               & Color \& Texture Consistency  & \begin{tabular}[c]{@{}l@{}}The color and texture of an object's surface should not undergo sudden, \\ unexplained changes.\end{tabular}                                & \begin{tabular}[c]{@{}c@{}}Object color flickering\\ Unstable textures\end{tabular}  \\ \cline{2-4} 
                                               & Static Semantic Consistency   & The visual content remains consistent with the text or other conditions.                                                                                               & Semantic inconsistency                                                               \\ \hline
\multirow{4}{*}{\textbf{Temporal Consistency}} & Motion Smoothness             & \begin{tabular}[c]{@{}l@{}}The motion trajectory of the object conforms to physical laws, with no \\ abrupt acceleration or deceleration.\end{tabular}                 & \begin{tabular}[c]{@{}c@{}}Object teleportation\\  Unnatural jumping\end{tabular}    \\ \cline{2-4} 
                                               & Coherent Temporal Dynamics    & \begin{tabular}[c]{@{}l@{}}State transitions between adjacent units appear visually smooth with \\ their trajectories adhering to real-world constraints.\end{tabular} & \begin{tabular}[c]{@{}c@{}}Image flickering\\ Action sequence violation\end{tabular} \\ \cline{2-4} 
                                               & Flicker Suppression           & \begin{tabular}[c]{@{}l@{}}Low-level features such as brightness and color exhibit no high-frequ-\\ ency oscillations over time.\end{tabular}                          & high-frequency noise                                                                 \\ \cline{2-4} 
                                               & Dynamic Semantic Consistency  & Sequence evolution follows the control intent of the prompt.                                                                                                           & Semantic inconsistency in actions                                                    \\ \hline
\end{tabular}}

\caption{Classification of Spatiotemporal Consistency}
\label{tab:Classification of Spatiotemporal Consistency}
\vspace{-0.75cm}
\end{table}

          In this survey, we comprehensively review the works that aimed to maintain spatiotemporal consistency in various aspects, including Generation Models~\cite{melnik2024video}, Feature Representations~\cite{yang2023diffusion,cho2024sora}, Generation Frameworks~\cite{liu2024sora}, Post-processing Techniques~\cite{liu2024sora}, Training Methods~\cite{singh2023survey} and Benchmarks and Evaluation Metrics~\cite{li2024survey}. Specifically, Section \ref{Sec:Generation Models} introduces the fundamental principles of mainstream generation models. Section \ref{Sec:Feature Representations} systematically explains how to construct the latent spatiotemporal feature distributions from video data. Section \ref{Sec:Generation Frameworks} summarizes the sampling strategies generation frameworks for spatiotemporal video sequences. Section \ref{Sec:Post-processing Techniques} discusses how to perform fine-grained optimization on generated spatiotemporal sequences through post-processing techniques. Section \ref{Sec:Training Strategies} outlines a series of effective training methods to enhance the spatiotemporal modeling capabilities. Section \ref{Sec:Benchmarks and Evaluation Metrics} summarizes relevant benchmark and evaluation metrics, which validate and quantify the quality and consistency of generated content. Section \ref{Sec:Future Directions and Challenges} delivers the future directions and challenges in spatiotemporal consistency video generation. The structure of this survey is shown in Figure \ref{Fig:Overview Structure Diagram}.

         \paragraph{\bf Contributions:} The primary contributions of this survey are threefold: (1) We frame video generation as a process of sequential sampling from high-dimensional spatiotemporal distributions and review research progress from the perspective of spatiotemporal consistency, distinguishing from existing surveys on video generation~\cite{li2024survey,liu2024sora,xiong2024autoregressive}; (2) We comprehensively summarize current research progress in video generation and elucidate its contributions to maintaining spatiotemporal consistency; (3) We explore future development opportunities and challenges in spatiotemporal consistency within this field. Relevant references are detailed in Table \ref{tab:Summary of Video Works}. We hope this survey provides valuable insights for advancing future video generation technologies.


\section{Generation Models}
\label{Sec:Generation Models}
        This section provides an overview of several common generation models \cite{hong2022cogvideo,xiong2024autoregressive} and introduces their contributions to maintaining spatiotemporal consistency. Specifically covered are Variational Autoencoder (VAE), Autoregressive Model (AR), Diffusion Model (DM), and Flow Model (FM).
        \begin{table}[]
        \centering
        \scalebox{0.7}{
        \begin{tabular}{c|c|c|c|c}
\hline
\textbf{Comparison Dimensions}      & \textbf{Variational Autoencoder}                                                 & \textbf{Autoregressive Model}                                                                      & \textbf{Diffusion Model}                                                                                            & \textbf{Flow Model}                                                                                                             \\ \hline
\textbf{Fundamental Principles}     & Variational Inference                                                            & Sequence Probability Modeling                                                                      & Iterative Denoising                                                                                                 & Revertable transformation                                                                                                       \\ \hline
\textbf{Spatiotemporal Consistency} & Poor                                                                             & Strong                                                                                             & Strong                                                                                                              & Moderate                                                                                                                        \\ \hline
\textbf{Long-term Consistency}      & Poor                                                                             & Strong                                                                                             & Strong                                                                                                              & Moderate                                                                                                                        \\ \hline
\textbf{Generation Quality}         & Moderate                                                                         & High                                                                                               & High                                                                                                                & High                                                                                                                            \\ \hline
\textbf{Controllability}            & Low                                                                              & Moderate                                                                                           & High                                                                                                                & High                                                                                                                            \\ \hline
\textbf{Training Stability}         & Poor                                                                             & High                                                                                               & High                                                                                                                & High                                                                                                                            \\ \hline
\textbf{Training Data Requirements} & Moderate                                                                         & High                                                                                               & High                                                                                                                & Moderate                                                                                                                        \\ \hline
\textbf{Inference Speed}            & Fast                                                                             & Slow                                                                                               & Slow                                                                                                                & Fast                                                                                                                            \\ \hline
\textbf{Memory Requirements}        & Low                                                                              & High                                                                                               & High                                                                                                                & Moderate                                                                                                                        \\ \hline
\textbf{Disadvantages}              & \begin{tabular}[c]{@{}c@{}}Blurry generated images\\ Lack diversity\end{tabular} & \begin{tabular}[c]{@{}c@{}}Slow reasoning\\ Cumulative errors\\ Difficult to parallel\end{tabular} & \begin{tabular}[c]{@{}c@{}}Slow inference speed\\ High computational cost\\ Poor real-time performance\end{tabular} & \begin{tabular}[c]{@{}c@{}}Complex structural design \\ Limited expressive capabilities\\ High training difficulty\end{tabular} \\ \hline
\end{tabular}}
        
            \caption{Comparison of Generation Models}
            \label{tab:Comparison of Generation Models}
            \vspace{-0.2cm}
        \end{table}

        \subsection{Variational Autoencoder (VAE)}
        VAE~\cite{kingma2013auto} is a classic generation model that performs generation inference based on Variational Bayesian networks. Unlike traditional autoencoders that describe latent space numerically, it describes the latent space in the form of probability distribution~\cite{li2024survey,liu2024sora}. VAE is regarded as one of the most valuable research methods in the field of unsupervised learning, with extensive applications particularly in feature representation for visual data such as images and video data~\cite{chen2024od}. VAE is based on the Gaussian mixture model and regards a complex data distribution $p(x)$ as the superposition of several Gaussian distributions $p(x|z)$, where $p(z)$ obeys a known standard Gaussian distribution. The formula is expressed as
        \begin{equation}
            	p(x)=\int_{z}p(z)p(x|z)dz.
        \end{equation} \par

        However, in practical applications, this model is generally not employed directly as an end-to-end video generator. This limitation stems primarily from its challenges in achieving high generation quality and maintaining stability during the training process. Instead, it excels at the extraction, compression, and reconstruction of video features, effectively learning a compact and structured latent representation of the high-dimensional spatiotemporal data~\cite{lopez2018information}. Recognizing this utility, a significant trend is to prioritize the design and training of an efficient, task-specific VAE as a foundational step~\cite{openai2024sora,kong2024hunyuanvideo,wan2025wan}. The learned spatiotemporal feature space subsequently serves as a core representation to support generation inference in powerful generation models such as autoregressive and diffusion models. The content regarding spatiotemporal feature representation will be covered in Section \ref{Sec:Feature Representations}.

        \subsection{Autoregressive Model (AR)}
            Autoregressive model is a classic generation model that employs the Maximum Likelihood Estimation method (MLE)~\cite{vaswaniAttentionAllYou2017a,dosovitskiy2020image}. According to the previous description, if video frames or local spatiotemporal patches are regarded as a spatiotemporal units, based on the mathematical principles of autoregressive models, it will generate the current frame conditioned on preceding frames at the frame level, while generating the current patch conditioned on previous ones at the patch level, thereby achieving video generation~\cite{hong2022cogvideo,xiong2024autoregressive}. The formulation is listed below.
            \begin{equation}
            	p(X)=p(x_0)\prod_{i=1}^L p(x_i \vert x_{<i};\theta),
            \end{equation}

            Autoregressive models~\cite{zhou2024survey,wu2022nuwaa,li2024survey} possess inherent advantages in video generation due to their innate ability to model sequences. They treat video generation as an incremental prediction process, referencing the previously generated sequence when producing the current unit. This model naturally captures temporal dependencies and maintains chronological consistency. Simultaneously, the incremental generation mechanism ensures the fundamental quality of each unit. Consequently, autoregressive models can address spatial and temporal consistency, enabling high-quality video generation with natural sequencing.

        \subsection{Diffusion Model (DM)}
            The diffusion model was initially used in the image generation domain and later transferred into video generation field~\cite{yang2023diffusion,xing2024survey,wang2025surveyvideodiffusionmodels}. The diffusion model defines a Markov chain of diffusion steps, where random noise is gradually added to the original data distribution until it becomes pure Gaussian noise~\cite{singh2023survey}. Then, the model learns the reverse denoising process to restore the original data distribution. The model is optimized by minimizing the distance between the real noise $\epsilon$ and the predicted noise $\epsilon_{\theta}$.
            \begin{equation}
            	L(\theta)=\Bbb{E}_{x,\epsilon,t}  
            	\left\|\epsilon-\epsilon_{\theta}(x,t)\right\|^2.
            \end{equation} \par
            
            
        
            Some methods have been proposed to enhance the spatiotemporal consistency of the generated videos. For example, the Latent Diffusion Model~\cite{he2022latent} maps the original data into a latent feature space, enabling a more efficient spatiotemporal representation. Moreover, noise is a crucial component in diffusion models, which determines the generated content. Choosing random noise can easily lead to suboptimal results. Blattmann et al.~\shortcite{blattmann2023align} addressed this by aligning the diffusion model’s upsamplers in the temporal domain, transforming them into temporally consistent super-resolution models. The noise estimation network (usually U-Net network) determines the model's denoising capability and generation quality. Peebles and Xie~\shortcite{peebles2023scalable} replaced the U-Net backbone with a transformer (DiT), showing impressive capabilities in generating high-quality video.

        \subsection{Flow Model (FM)}
            Flow models~\cite{ho2019flow++,lipmanflow} constitute a class of generation models, whose core lies in explicitly modeling a bidirectionally reversible mapping from simple prior distributions to complex data distributions through a set of invertible transformation functions. Unlike diffusion models or autoregressive models, flow models possess a fully reversible structure, enabling both precise likelihood computation and efficient few-step sampling. The model is optimized by minimizing the flow matching loss. Recently, this model has also been extensively applied to video generation.
             \begin{equation}
            	L(\theta)=\Bbb{E}_{x,t}  
            	\left\|v_{\theta}(x,t) - (x_1-x_0)\right\|^2.
            \end{equation} 
            Its inherent reversible transformation structure provides a theoretical guarantee for spatiotemporal consistency in video generation: by constructing smooth trajectories in the feature space, the model can map these into smooth evolutions between spatiotemporal units within the video, thereby mathematically and explicitly preserving temporal coherence and spatial stability. Today, flow models have been successfully adopted and extensively explored in numerous cutting-edge works within the field of video generation~\cite{liu2024sora,li2024survey}. Pyramid-Flow~\cite{lei2023pyramidflow} proposes a ‘feature pyramid + flow matching’ architecture, which significantly enhances training stability and overall video quality by modeling continuous flows across multi-scale feature spaces. FlashVideo~\cite{zhang2025flashvideo} designs a near-linear, efficient sampling trajectory that smoothly maps from low-quality video to high-fidelity outputs, effectively integrating rich visual details and texture information while maintaining spatiotemporal consistency.

            \subsection{Summary}
            By reviewing and comparing four generation models, their positioning and evolutionary trajectories within video generation can be clearly delineated. Comparison is listed in Table \ref{tab:Comparison of Generation Models}. Theoretically, VAEs, autoregressive models, diffusion models, and flow models respectively correspond to four technical pathways: variational inference, sequence modeling, iterative denoising, and reversible transformations. When addressing spatiotemporal consistency challenges, the models exhibit marked performance disparities: autoregressive models, relying on causal modeling, theoretically offer robust consistency guarantees; diffusion models achieve current state-of-the-art results in practical applications through iterative global optimization; whereas VAEs and flow models, constrained by their respective architectures, demonstrate weaker consistency performance.  Specifically, VAEs often suffer from training instability leading to generation distortion, while flow models, though capable of achieving local smoothness through reversible transformations, struggle to model complex long-term spatiotemporal dependencies.

\section{Feature Representations}
 \label{Sec:Feature Representations}
 
        A feature representation space~\cite{kong2024hunyuanvideo,yang2024cogvideox} capable of effectively capturing the intrinsic spatiotemporal distribution structure and supporting high-quality sampling, which is the core foundation of video generation. To preserve the spatiotemporal consistency, current research advances feature representation techniques primarily through five directions: Compression Representation~\cite{chen2024od}, Long Sequence Representation~\cite{li2024survey}, Feature Discretisation~\cite{liu2025coda}, Feature Decoupling~\cite{wu2025improved}, and Feature Representation based on Diffusion Timesteps~\cite{pan2025generative}. These methods provide a more robust and computationally feasible basis for feature representation, enabling efficient and stable sampling of spatiotemporal unit sequences from complex spatiotemporal distributions.

      \begin{figure*}[t!]
            \centering
            \includegraphics[width=\textwidth]{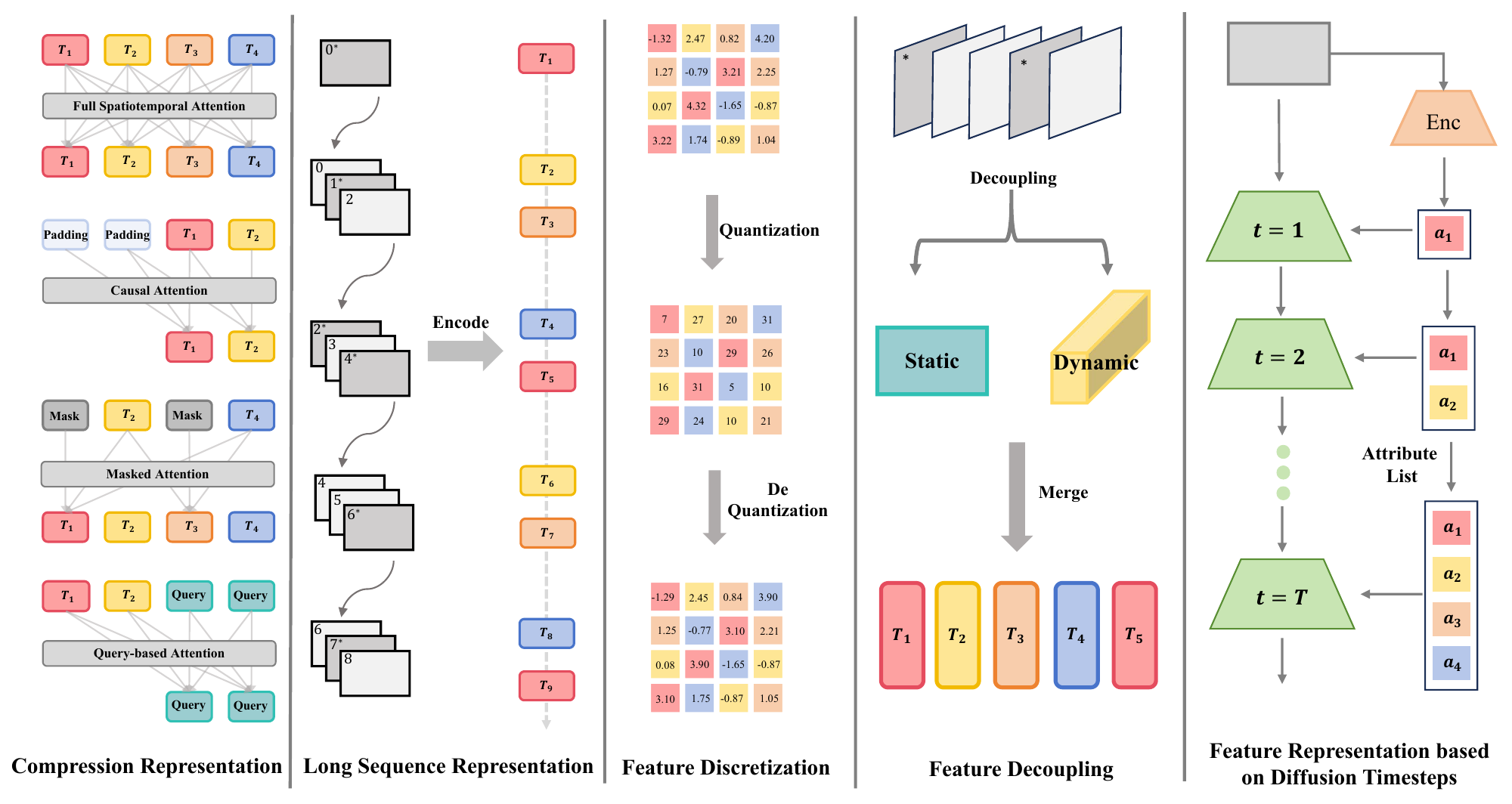}
            \caption{Feature Representations.}
            \label{Fig:Feature Representations}
            \vspace{-0.4cm}
        \end{figure*}
        
     \subsection{Compression Representation}
     Compressed representation~\cite{chen2024od,cho2024sora} effectively reduces data redundancy by compressing video data across spatial and temporal dimensions, whilst maintaining visual fidelity and temporal consistency. This approach addresses the high-dimensional complexity of video data, significantly lowering storage and computational costs, whilst ensuring the generated video meets high standards of visual fidelity and consistency across spatiotemporal units.

     \paragraph{\bf Spatiotemporal Representation.} This method performs a full-scale representation of space and time~\cite{melnik2024video}. 3D-VAE~\cite{liu2024sora,xing2024survey} expands the traditional 2D-VAE architecture, incorporates the time dimension into the encoding process, and uses 3D CNN or attention mechanism to capture inter-frame dependencies. This method can compress videos in both spatial and temporal dimensions, significantly improving the compression rate while maintaining high-fidelity video reconstruction. 
     Chen et al.~\cite{chen2024od} proposed a VAE framework OD-VAE which is specially designed for all-round video compression. By jointly modeling spatial and temporal features, static content and dynamic changes in the video are captured, thus achieving a balance between compression rate and reconstruction quality. 
     
     \paragraph{\bf Causal Representation.} Causality is one of the properties of video-unit sequences~\cite{xing2024survey}. CogVideoX~\cite{yang2024cogvideox} and HunyuanVideo~\cite{kong2024hunyuanvideo} use a 3D Causal VAE based spatiotemporal representation. Compared with 3D VAE, 3D Causal VAE emphasizes the causal relationship in the time dimension, ensures the one-way transmission of information between frames~\cite{wan2025wan}. During the encoding process, the model only uses the current frame and previous frame information for feature extraction and compression, preventing the leakage of temporal information and ensuring that the generated video maintains coherence in dynamic changes, thereby improving temporal consistency.

     \paragraph{\bf Masked Representation.} Masked methods enhance the model's spatiotemporal understanding and modeling capabilities by randomly masking portions of spatiotemporal units, thereby compelling the model to infer masked information from contextual cues. VideoMAE~\cite{tong2022videomae,wang2023videomae} adopts the "mask + reconstruction" paradigm and transforms video reconstruction into a self-supervised task through a high-ratio video mask. MotionAura~\cite{susladkar2024motionaura} uses a novel 3D Mobile Inverse Vector Quantization Variational Autoencoder (3D-MBQ-VAE) to simulate the dynamic changes of video frames by introducing a masked labeling model. Through a full-masking strategy, the model is forced to infer missing information from context, thereby enhancing the ability to model dynamic relationships between frames.
     
     \paragraph{\bf Learnable Representation.} Drawing on the BLIP-2 model~\cite{li2023blip}, learnable query is introduced to obtain global spatiotemporal information. Traditional video tokenizers~\cite{he2022latent,esser2023structure} usually use block tokenization to segment video frames into local visual blocks and then perform feature encoding. LARP~\cite{wang2024larp} introduces a holistic tokenization scheme to extract information from videos through learnable queries. The design of global queries allows the model to comprehensively consider information in spatial and temporal dimensions, capturing the overall semantics (such as scene structure) and dynamic changes (such as object motion trajectory) of the video.
     
     
     \subsection{Long Sequence Representation}
     Long sequence representation~\cite{villegas2022phenaki,yang2025rethinking,elmoghany2025surveylongvideostorytellinggeneration} is a key technology in the field of video generation, focusing on generating long, dynamically coherent video sequences. By reducing the number of visual tokens~\cite{li2024dicode}, optimizing the latent representation structure, and introducing efficient representation mechanisms~\cite{qin2024xgen,lu2024freelong} and feature cache~\cite{singer2022make}, these methods significantly reduce computational and storage costs while ensuring that the generated videos meet high standards in visual quality and inter-frame coherence.

     \paragraph{\bf Divide and Merge.} Divide-and-merge~\cite{li2024survey} is a commonly used method for processing long videos. To maintain temporal consistency in long video generation, xGen-VideoSyn-1~\cite{qin2024xgen} introduces a divide-and-merge strategy. This strategy divides long video sequences into multiple short clips, each of which is processed independently, and then these clips are integrated into a complete one through temporal fusion mechanisms. This fragmentation process effectively avoids the cumulative error problem~\cite{ai2025magi1autoregressivevideogeneration}. At the same time, through joint spatiotemporal compression, a compact potential representation is generated, which significantly reduces the number of tokens.

     \paragraph{\bf Global and Local Fusion.}
     In long-video generation tasks, the effective fusion of global and local information~\cite{liu2024sora} preserves the coherence of global semantic structures and the natural transition of local dynamic details, achieving spatiotemporal consistency. DiCoDe~\cite{li2024dicode} employs high-frame-rate sampling and deep-token compression to reduce representation size while preserving critical details and motion, providing an efficient feature foundation for spatiotemporal modeling of long sequences. LongDiff~\cite{li2025longdiff} further proposes a dynamic frame selection mechanism based on pseudo-video and keyframe detection, enabling the model to adaptively focus on crucial temporal segments during generation. This effectively mitigates the dilution of attention weights caused by excessive frame counts. FreeLong\cite{lu2024freelong} introduces a spectralblend temporal attention mechanism that integrates global low-frequency components with local high-frequency components in the frequency domain, balancing semantic consistency with local high fidelity.

     \paragraph{\bf Feature Cache.} The feature cache~\cite{wan2025wan} avoids repeated calculations by storing intermediate features, significantly reducing the computational cost of long sequence generation. FasterCache~\cite{lvfastercache} introduces a dynamic feature reuse strategy for attention modules, preserving both feature distinctiveness and continuity between adjacent time steps while maintaining the necessary gradual changes. This effectively balances efficiency with temporal consistency. ToCa~\cite{zouaccelerating} employs an adaptive token-level cache reuse strategy, significantly reducing the computational overhead of generating long videos while preserving sequence coherence across the temporal dimension. Yang et al.~\shortcite{yang2025rethinking} proposes a video representation and generation framework based on a 3D causal diffusion model and feature caching mechanism. This approach demonstrates outstanding performance in high-fidelity reconstruction and temporal consistency, supporting practical applications such as long video generation and video continuation~\cite{singer2022make}.

     

     \subsection{Feature Discretization}
     Feature discretization~\cite{li2024survey} significantly reduces data dimensionality by discretizing continuous visual representations into compact tokens. This approach transforms video generation into a sequence prediction task~\cite{xiong2024autoregressive}. Feature discretization technology addresses the challenges of high complexity and spatiotemporal consistency in video generation by token quantification~\cite{wang2025bridging}, optimizing token representation~\cite{yu2023spae}, and introducing efficient prediction mechanisms~\cite{guo2025improving}.

     \paragraph{\bf Token Quantification.} 
     Quantization techniques~\cite{xiong2024autoregressive,xing2024survey} convert feature representations into discrete tokens. CODA~\cite{liu2025coda} proposes a framework for decoupling compression and quantification, which converts the potential representation of continuous VAE into a discrete token sequence, avoiding the complexity of training process and achieving efficient feature representation. Effective discrete token~\cite{zhou2024survey} can capture intra-frame details and inter-frame dynamics, ensuring high fidelity and temporal consistency of the generated video. TokenBridge~\cite{wang2025bridging} optimizes the tokenizer training strategy and proposes a method to bridge continuous and discrete tokens through post-training quantization. It introduces a dimension-based quantization method to discretize each feature dimension independently and generate a compact discrete token sequence.
     
     \paragraph{\bf Token Optimization.}
     Optimizing tokens to improve their information representation capabilities. SPAE~\cite{yu2023spae} proposes a token optimization method that bridges the pixels space with the large language model vocabulary, performs multi-scale feature extraction and layer-by-layer quantization, and gradually converts the visual input into compact lexical tokens. These tokens not only capture the fine-grained details required for visual reconstruction, but also have semantic interpretability.

     \paragraph{\bf Token Prediction Mechanism.}
     An effective token prediction mechanism can better capture spatiotemporal information~\cite{li2024survey,cho2024sora}. Guo et al.~\shortcite{guo2025improving} proposed a two-stage framework, implementing a coarse-to-fine token prediction strategy to improve the efficiency of feature quantization. Among them, coarse token prediction uses an AR model to capture the global structure and information, while fine token prediction introduces an auxiliary model to simultaneously predict fine-grained tokens. This method can support long sequence video generation while maintaining high fidelity and temporal consistency.


     \subsection{Feature Decoupling}
     Feature decoupling~\cite{yang2023diffusion} solve the challenges of high computational complexity and temporal consistency by decomposing video content into different components and optimizing compression efficiency. These methods generate compact and information-rich latent representations by decoupling the temporal and spatial features, static content, and dynamic information of videos~\cite{qing2024hierarchical}, while ensuring high-quality representation of details and dynamics of generated videos.
     
      \paragraph{\bf Time-Space Decoupling.}
      Time-space decoupling is a commonly employed decoupling strategy~\cite{sun2024sora}. OmniTokenizer~\cite{wang2024omnitokenizer} integrates window attention and causal attention to optimize spatiotemporal modeling respectively, decomposing videos into spatiotemporal components to generate compact latent representations that support efficient compression and high-fidelity reconstruction. SweetTok~\cite{tan2024sweettokenizer} proposes a decoupled spatiotemporal query framework, compressing visual inputs through independent spatiotemporal queries while employing MLC to address semantic discrepancies between appearance and motion information. Zuo et al.~\shortcite{zuo2024edit} adopt a two-stage learning strategy for spatiotemporal decoupling, enabling the model to capture key video information more efficiently by decoupling the learning processes of temporal and spatial features.

     \paragraph{\bf Content-Motion Decoupling.}
     Content-motion decoupling decouples and analyzes video features from a semantic perspective~\cite{sun2024sora}. Traditional methods directly apply image VAE to video frames, which may lead to temporal inconsistency, while simply extending it to 3D-VAE may introduce motion blur or detail distortion. Xing et al.~\shortcite{xing2024large} proposed a feature decomposition method to decompose video into static content and dynamic information through a time-aware spatial compression model and a motion compression model. IV-VAE~\cite{wu2025improved} proposes a keyframe-based temporal compression architecture and a group causal convolution module, which divides the latent space into two branches: half of the branches focus on capturing static content, the other branches perform temporal compression to capture dynamic information. VidTwin~\cite{wang2025vidtwin} decomposes video into a structural latent vector and a dynamic latent vector to achieve compact feature representation. WF-VAE~\cite{li2025wf} decomposes the video into low-frequency components (capturing global structure and content) and high-frequency components (capturing details and rapid changes) through multi-level wavelet transform, and encodes and compresses them separately.


     \subsection{Feature Representation based on Diffusion Timesteps}
    The diffusion model generates high-quality samples by gradually recovering data from noise in multiple timesteps~\cite{yang2023diffusion,melnik2024video}. The feature representation based on diffusion timesteps uses this iterative process to gradually refine visual data into structured, recursive tokens. These tokens capture features ranging from low-level details to high-level semantics in each timestep~\cite{xing2024survey}.

    Existing methods rely on spatial visual tokens~\cite{dosovitskiy2020image,kong2024hunyuanvideo,wan2025wan}, where image patches are encoded and arranged according to spatial order. However, spatial tokens lack the recursive structure inherent in language and thus constitute a "language" that is difficult for large language models to master. Pan et al.~\cite{pan2025generative} proposed to use diffuse timesteps to learn discrete, recursive visual tokens to build a suitable visual language. These tokens recursively compensate for the properties gradually lost in noisy frames as timesteps increase, enabling the diffusion model to reconstruct the original video frame at any timestep. Lin et al.~\cite{lin2025reasoning} proposed the Phys-AR framework to train the Diffused Timestep Tokenizer (DDT), which consists of two stages: the first stage uses supervised fine-tuning to transfer symbolic knowledge, while the second applies reinforcement learning to optimize reasoning ability through a physically based reward function. 
     

     \subsection{Summary}
     In summary, the construction of spatiotemporal feature spaces constitutes the fundamental basis for achieving high-quality spatiotemporal consistency in video generation tasks. The aforementioned methods not only enhance the model's ability to capture the intrinsic structure of complex feature distributions but also provide a viable technical pathway for efficient and stable spatiotemporal sequence units sampling. Moving forward, constructing more structured and interpretable multi-level spatiotemporal feature representation spaces remains a key research direction for realizing long-duration video generation and refined dynamic control.

\section{Generation Frameworks}
\label{Sec:Generation Frameworks}

     Within the domain of video generation, generation frameworks\cite{liu2024sora,li2024survey} serve as pivotal structures linking foundation generation models to video generation tasks. From the perspective of high-dimensional spatiotemporal distribution sampling, diverse frameworks essentially provide distinct organizational paradigms for extracting coherent sequences from complex distributions~\cite{sun2024sora}. Diffusion generation framework explicitly optimizes joint spatiotemporal distributions through multi-step denoising, incorporating temporal smoothness into sampling~\cite{yang2023diffusion}; Autoregressive generation framework transition probabilities through sequential conditional sampling, reinforcing inter-frame dynamic coherence~\cite{xiong2024autoregressive}; Conditional generation framework guides sampling using global or local information, constraining semantics and structure to maintain spatial consistency~\cite{chen2024controlavideocontrollabletexttovideodiffusion}; Multi-stage generation framework decouples the process by first learning distributional structure and subsequently optimising temporal connections, treating spatial semantics and temporal evolution separately~\cite{zhang2023i2vgen}; Interactive generation framework introduces external feedback mechanisms, dynamically adjusting sampling trajectories to enhance long-range control capabilities~\cite{che2024gamegen}.

    \subsection{Diffusion Generation Framework}
     The diffusion generation framework~\cite{melnik2024video,wang2025surveyvideodiffusionmodels} represents an advanced video generation approach. It progressively iterates and optimizes from a sequence of spatiotemporal units initialized from random noise, thereby generating video sequences with high spatiotemporal consistency~\cite{ho2020denoising,nichol2021improved}. By balancing spatial detail with temporal continuity, this approach ensures generated videos meet high standards in both visual fidelity and motion smoothness. The following sections elaborate on the core techniques and practical applications of the diffusion generation framework across four dimensions: noise initialization~\cite{wu2024freeinit}, latent space denoising~\cite{hacohen2024ltx}, noise prediction and scheduling~\cite{wang2024cono}, and spatiotemporal modeling~\cite{bar2024lumiere}. \par
     

     \begin{figure*}[t!]
            \centering
            \includegraphics[width=\textwidth]{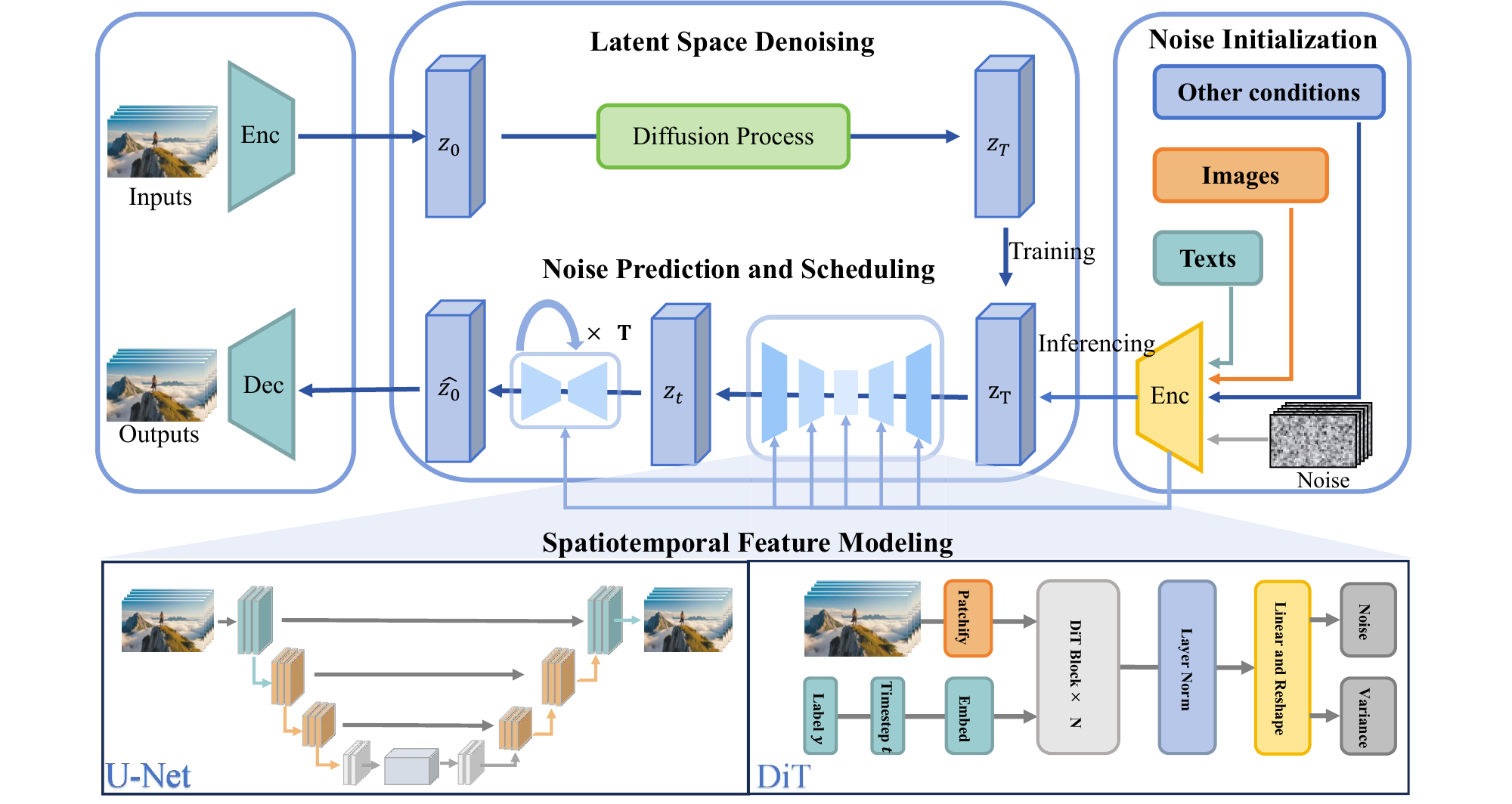}
            \caption{Diffusion Generation Framework.}
            \label{Fig:Diffusion Generation Framework}
            \vspace{-0.5cm}
        \end{figure*}

     \paragraph{\bf Noise Sequence Initialization.}
     Noise Sequence initialization~\cite{xing2024survey} constitutes the pivotal starting point within diffusion generation framework. This methodology generates a sequence composed of randomly initialized video units (typically regarded as noise sequences), thereby furnishing diverse initial states for subsequent denoising optimization processes~\cite{liu2024sora}. Gaussian sampling, as the most prevalent initialization technique, leverages its statistical properties to ensure generated content possesses sufficient diversity and creativity. Appropriate noise initialization not only determines the starting distribution for sampling but also indirectly influences the convergence efficiency of the denoising path and the visual quality of the generated results. It represents a crucial design element in balancing generation diversity with temporal stability.

    Qiu et al.~\shortcite{qiufreenoise} found that per-frame noise forms the foundation determining a video's overall appearance, while temporal order influences generated content. They proposed FreeNoise, which achieves noise sequences exhibiting intrinsic randomness and long-range correlations through local denoising and window-based attention fusion, enabling effective handling of long-video generation. Wu et al.~\shortcite{wu2024freeinit} found that there is an implicit "noise gap" in the training-inference process, meanwhile, the low-frequency components of the initial noise will significantly affect the denoising process. He proposes FreeInit~\cite{wu2024freeinit}, through iteratively optimizing the spatiotemporal low-frequency components of the initial noise during the inference process, effectively improving the main appearance and temporal consistency of the generated results. Zhou et al.~\shortcite{zhou2024golden} injected the semantic information of the input prompt into the original sampled noise, and used the noise containing the semantic information to improve the generation quality. Subsequent studies further studied the relationship between the denoising process and the inversion process~\cite{Wallace_2023_CVPR}. Bai et al.~\shortcite{bai2024zigzag} found that the difference in the guidance strength in this two processes can capture the semantic information in the latent space, which has an important impact on the generation quality and conditional alignment. They propose Z-Sampling~\cite{bai2024zigzag}, a method that allows the diffusion model to improve the generation quality through step-by-step denoising self-reflection.

    \paragraph{\bf Latent Space Denoising.}
    Latent space refers to the feature space that compresses high-dimensional spatiotemporal data into low-dimensional representation through dimensionality reduction coding technology~\cite{he2022latent, cheng2025phasedonestepadversarialequilibrium}. The significance of latent space denoising is to gradually transform the initial random noise sequence into video sequence with visual coherence and temporal consistency~\cite{melnik2024video}. At present, latent space denoising technology is used in many outstanding works.
    
    Blattmann~\shortcite{blattmann2023align} applied the latent diffusion model (LDM) paradigm to high-resolution video generation. It introduces the time dimension based on the pre-trained image LDM to fine-tuned the video LDM, and achieved high-quality video generation. LTX-Video~\cite{hacohen2024ltx} is a Transformer-based LDM that compresses video data into the latent space through a Video-VAE with a high compression ratio. Ni et al.~\shortcite{ni2023conditional} combined the latent model with the flow model and proposed the LFDM method to synthesize optical flow sequences in the latent space according to given conditions to warp a given image. Spatiotemporal consistency is also a major challenge in the field of video synthesis. Wang et al.~\shortcite{wang2024leo} proposed LEO, which realizes this idea through a flow-based image animator and a latent motion diffusion model. The core idea is to represent motion as a series of flow graphs during the generation process, thereby separating motion from appearance. 

    \paragraph{\bf Noise Prediction and Scheduling.}
    Noise prediction and scheduling~\cite{liu2024sora,xing2024survey} are key links in optimizing denoising efficiency and quality in the diffusion framework. The noise prediction model estimates the noise component in each step through a deep neural network, guiding the model to gradually approach the real video sequence~\cite{ho2020denoising}. The scheduling strategy is responsible for controlling the rhythm and step size of denoising~\cite{nichol2021improved}. By reasonably arranging the denoising intensity of each iteration, the generation process is ensured to be both efficient and stable. The adaptive scheduling strategy is an important progress in recent years~\cite{sun2024sora}, which can dynamically adjust the denoising speed according to the complexity of the video content.

    In the technical report of ModelScope~\cite{wang2023modelscope}, the model optimizes the noise prediction process by combining the variational quantization network VQGAN, text encoder and denoising U-Net to ensure the consistency of frame generation and the smoothness of motion transition. Its scheduling strategy balances the computational cost and generation quality through pre-trained step size adjustment. Traditional diffusion models usually add noise to each unit in a video sequence independently~\cite{chen2024diffusion}, ignoring the content redundancy and temporal correlation between units, resulting in unnatural transitions in the generated video. VideoFusion~\cite{luo2023videofusion} uses a decomposition diffusion process to improve the accuracy of noise prediction by decomposing the noise into a base noise shared by all frames and a residual noise that varies along the time axis. Two jointly learned networks are applied to process the base noise and residual noise respectively, ensuring that the generated frames are consistent in spatial details and temporal dynamics. CoNo~\cite{wang2024cono} focuses on the fine-grained consistency problem in long video generation and proposes a consistent noise injection method. It optimizes noise scheduling through a "look-back" mechanism, decomposing the denoising process into three parts, which allows the model to refer to the previous context when generating new units.

    \paragraph{\bf Spatiotemporal Feature Modeling.} 
    Spatiotemporal feature modeling~\cite{li2024survey} is the core technology in the diffusion generation framework to ensure that video maintain consistency in spatial details and temporal dynamics. The key is to capture the complex dependencies between the two-dimensional spatial features and one-dimensional temporal features of the video, thereby avoiding visual artifacts or temporal discontinuities in the generation process~\cite{zhou2024survey,cho2024sora}.

    Introducing a temporal layer into traditional image space modeling is a common spatiotemporal modeling method~\cite{cai2024real}. Stable Video Diffusion~\cite{blattmann2023stablevideodiffusionscaling} adds a temporal layer to the U-Net and VAE, including a 3D attention layer and a spatiotemporal attention layer. Spatial attention focuses on the interaction of information within the same frame, while temporal attention focuses on the interaction of information between frames. Lumiere~\cite{bar2024lumiere} adds a temporal downsampling and an upsampling layer to the U-Net, which can compress information in space and time, and improve the temporal consistency of the generation process.   
    
    Recently, some works~\cite{li2024survey} have combined the diffusion model with the transformer to propose the DiT framework. Video data is divided into independent non-overlapping spatial patches or spatiotemporal patches~\cite{wang2024patch}, and is marked using spatiotemporal position encoding, so that video generation can be performed using sequence modeling~\cite{dosovitskiy2020image}. Among them, different attention mechanisms are proposed to better solve the spatiotemporal consistency problem. Nuwa-XL~\cite{yin2023nuwa} uses temporal attention, focusing only on patches at the same position in different frames, while Make-A-Video~\cite{singer2022make} uses a full spatiotemporal attention mechanism, focusing on all other patches, however takes heavy computational cost. MagicVideo~\cite{zhou2023magicvideoefficientvideogeneration} optimizes this by using a causal attention mechanism, focusing only on the current patch and all previous patches. Tune-A-Video~\cite{wu2023tune} and Rerender-A-Video~\cite{yang2023rerender} further optimize computational efficiency, with each patch only focusing on the previous frame and all patches in the first frame.

    In addition, spatiotemporal decoupling is also an effective spatiotemporal modeling method. In CMD~\cite{yu2024efficient} and TF-T2V~\cite{wang2024recipe} work, the video is encoded as a combination of content and motion latent representations. The content frames are generated by fine-tuning the pre-trained image diffusion model, and the motion representation is generated by training a lightweight diffusion model. HiGen~\cite{qing2024hierarchical} improves performance by decoupling from structural level and content level. As for structural level, a unified denoiser is used to generate spatiotemporal priors, and for content level, flexible changes in content are achieved based on content clues.

               
        \subsection{Autoregressive Generation Framework}
        
         \begin{figure*}[t!]
        
            \centering
             \scalebox{1}{
            \includegraphics[width=\textwidth]{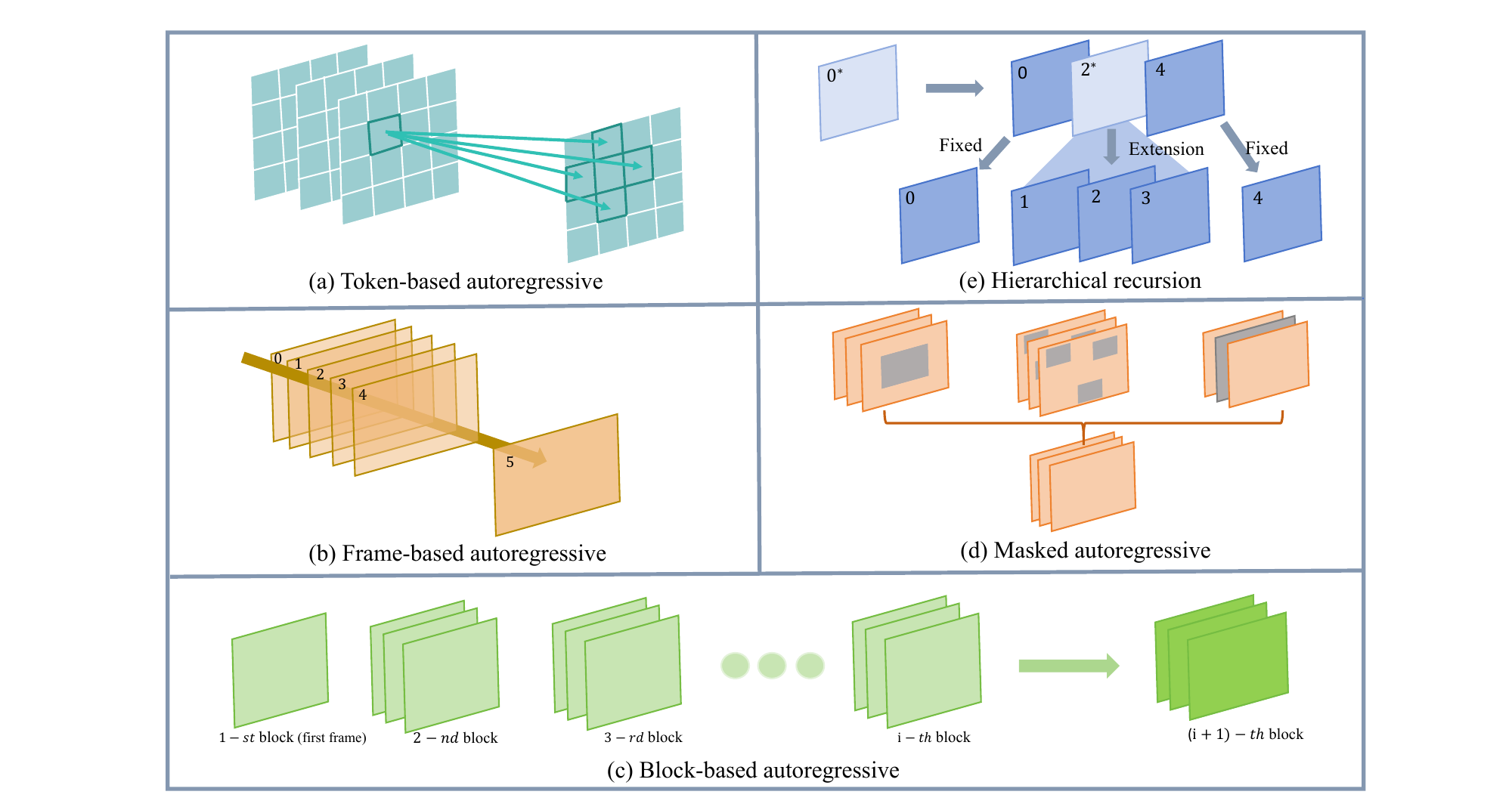}
            }
            \caption{Autoregressive Generation Framework.}
            \label{Fig:Autoregressive Generation Framework} 
            \vspace{-0.1cm}
        \end{figure*}

        Within autoregressive generation frameworks~\cite{xiong2024autoregressive,liu2024sora}, the video generation process is modeled as sequential conditional sampling from a high-dimensional spatiotemporal distribution. Specifically, at each generation step, the model samples the spatiotemporal units at the current time step based on the generated sequence context~\cite{hong2022cogvideo,wu2022nuwaa,li2024survey}. This mechanism inherently models the transformational dynamics of sequences explicitly, emphasizing dynamic coherence between adjacent units. Consequently, it progressively constructs temporally consistent video sequences during sampling. Through this sequential conditional sampling, the autoregressive framework effectively captures dynamic patterns and long-range dependencies in videos as they evolve over time, ensuring logical coherence and natural motion across the temporal dimension of generated content.

        
        \paragraph{\bf Token-based Autoregression.}
        The Token-based autoregressive~\cite{chen2024diffusion} method is a framework for generating videos by sequentially predicting discrete video tokens. Its core mechanism is to decompose the video into a series of token representations, each token representing a spatiotemporal feature unit in the video, and gradually predict subsequent tokens based on the conditional probability distribution of previous tokens~\cite{villegas2022phenaki}. The feature representation method proposed in Section \ref{Sec:Feature Representations} further enhances the capabilities of the token-based autoregressive framework by converting video data frames into discrete token representations, achieving accurate modeling and control of the spatiotemporal details of the video.

        Loong~\cite{wang2024loong} is a video generator based on autoregressive LLM. It models text tokens and video tokens as a unified sequence of autoregressive LLM, and can generate minute-long videos based on text prompts. Emu3~\cite{wang2024emu3} can tokenize images, text, and videos into discrete spaces, and generate high-fidelity videos by predicting the next token in a video sequence. Chen et al.~\shortcite{chen2024diffusion} proposed a new training paradigm called Diffusion Forcing. The training diffusion model can denoise a set of tokens with independent noise levels, and generate one or more future tokens through a causal next token prediction model. This method takes the advantages of the token prediction model that can achieve variable length generation.

        \paragraph{\bf Frame-based Autoregression.}
        Frame-based autoregression~\cite{xiong2024autoregressive} is a video generation technology that directly predicts a complete new frame based on the image information of the previous frames. Its core idea is to use the temporal dependency between frames in a video sequence to generate subsequent frames by learning the features of the previous frames, thereby maintaining the smoothness and consistency of the action.

        ART-V~\cite{weng2024art} is an effective frame autoregressive framework that generates the current frame by learning simple continuous motion between adjacent frames. It can generate videos of arbitrary length based on prompts such as text, images, or condition combination, making it very versatile and flexible. NOVA~\cite{deng2024autoregressive} proposes an autoregressive generation scheme that uses set-by-set prediction generation in a single frame and frame-by-frame prediction for the entire video sequence. This will allow NOVA to better handle temporal causality and spatial relationships.

        \paragraph{\bf Block-based Autoregression.}

  
        Modeling the entire video directly requires dealing with extremely high data dimensions, which poses a huge challenge to computing resources and training time. The block-based autoregressive method~\cite{cho2024sora,ren2025next,elmoghany2025surveylongvideostorytellinggeneration} simplifies this problem by decomposing the video into smaller units, called "blocks". Through this decomposition, the original generation task is transformed into a series of subtasks, reducing the task complexity.

        As far as we know, the idea of video block processing has been introduced by many works. Some works, such as Wan~\cite{wan2025wan} and Hunyuan~\cite{kong2024hunyuanvideo, wu2025hunyuanvideo15technicalreport}, adopt a block strategy when processing input video sequences, and each encoding and decoding operation only processes video blocks. According to the temporal compression ratio, each block contains a maximum of four frames, which effectively prevents memory overflow and supports infinite long videos as input. MAGI-1~\cite{ai2025magi1autoregressivevideogeneration} is the first work to generate video by autoregressively predicting a sequence of video blocks, where a video block is defined as a fixed-length segment of consecutive frames. It can progressively generate each video block, thereby enabling causal temporal modeling and naturally supporting streaming generation. Ren et al.~\shortcite{ren2025next} proposes “next block prediction” video generation paradigm, termed semi-autoregressive modeling. CausVid~\cite{yin2025slow} employs a mechanism combining block-level causal attention with bidirectional self-attention. Causal constraints maintain temporal logical coherence between frames, while bidirectional interactions enhance the consistency of spatial representations, thereby improving sampling efficiency and generation quality. Block-based autoregressive generation offers a promising direction for unified high-fidelity long video generation, flexible instruction control, and real-time deployment.
        

        \paragraph{\bf Masked Autoregression.}
        Masked autoregressive~\cite{yu2024language} is a technique in video generation that boosts flexibility by focusing on predicting the masked portions of video frames. Unlike traditional autoregressive methods, this approach selectively targets masked regions, conditioning predictions on visible parts. This makes it especially valuable for tasks like video restoration and completion, where missing or damaged content needs to be seamlessly reconstructed.
  
        MAGVIT~\cite{yu2023magvit,yu2024language} uses masked multi-task training for video generation tasks. It first quantizes the video into discrete tokens through a 3D VAE. In each training step, it randomly selects a task and its prompt tokens to obtain task-specific conditional masks, and optimizes the transformer to predict all target tokens. MarDini~\cite{liu2024mardini} integrates the advantages of masked autoregression into a unified diffusion model framework, where masked autoregression is responsible for temporal planning, while the diffusion model focuses on spatial generation. Mask prediction~\cite{liu2024sora} is superior to traditional autoregression in that it is bidirectionally parallelizable, effectively reducing time costs.

        \paragraph{\bf Hierarchical Recursion.}
        The hierarchical recursive framework~\cite{xiong2024autoregressive} is a powerful framework for video generation, leveraging multi-scale modeling to progressively refine video content from coarse outlines to fine details. By decomposing the complex task of generating high-dimensional video data into solvable subtasks~\cite{sun2024sora,xing2024survey}, the method optimizes computational efficiency. It recursively processes videos at different scales, ensuring coherence and details, making it a cornerstone of modern video generation frameworks.

        Wu et al.~\shortcite{wu2022nuwab} presented NUWA-Infinity, an autoregressive-over-autoregressive generation architecture. In this architecture, the global model considers the dependencies between patches, while the local model focuses on the dependencies of visual tokens within each patch. Yin et al.~\shortcite{yin2023nuwa} proposed NUWA-XL, which adopts a diffusion-over-diffusion architecture. The global model is first used to generate key frames, and then local model is applied iteratively to complete intermediate frames.

        \subsection{Conditional Generation Framework}
        \label{Conditional Generation Framework}
        Conditional generation framework~\cite{ni2023conditional,ma2025controllablevideogenerationsurvey} involves fine-grained guidance and creation of video content by incorporating external conditions such as text and images. Its core principle lies in the model's capability not only to generate semantically coherent content but also to ensure temporal and spatial consistency in the output videos~\cite{yang2023diffusion,singh2023survey}. This process relies heavily on the strict maintenance of spatiotemporal consistency.

     
        \paragraph{\bf Text-guided Generation.}
         Text-guided video generation~\cite{sun2025t2v, ju2025editverseunifyingimagevideo} uses natural language as condition to drive models in creating dynamic visual content. The generation process requires at two levels alignment: semantic alignment, where video frames and dynamic content match textual descriptions; and spatiotemporal alignment, where frames must maintain spatial consistency and ensure continuous, logical progression over time. This ensures the story described in text translates into visually fluid and consistent video content.

        
         Emu Video~\cite{girdhar2024emuvideofactorizingtexttovideo} and Factorized-Dreamer~\cite{yang2024factorizeddreamertraininghighqualityvideo} decompose the generation process into two steps: generating images conditioned on highly descriptive captions, and synthesizing videos conditioned on the generated images and detailed captions. They adopt a spatiotemporal decomposition framework for better understanding of text descriptions. Zhang et al.~\shortcite{zhang2025show1marryingpixellatent} proposed Show-1, combining pixel-based VDM and vector-based VDM for text-to-video generation. Among them, the pixel-based VDM generates low-resolution videos with strong text-video correlation. Afterwards, the low-resolution video is further upsampled to high resolution by vector-based VDM. In order to solve the problems of motion consistency, Chen et al.~\shortcite{chen2024controlavideocontrollabletexttovideodiffusion} further incorporated content priors and motion priors into the diffusion-based generation process, thereby improving the correlation between video frames and reducing flickering in the video.  For multi-textual conditional tasks,  Qiu et al.~\shortcite{qiufreenoise} proposed the Motion Injection method, leveraging the characteristic of diffusion models to restore information at different levels by denoising at distinct time steps. This enables the gradual injection of new motion at appropriate time timestep following the preceding action. The method ensures multi-prompt long-video generation while maintaining excellent spatiotemporal consistency.
        
        \paragraph{\bf Image-guided Generation.}
        Image-guided video generation~\cite{sun2024sora} imposes more explicit spatial constraints. It requires generated content to align with the visual spatial features of the input image. Dynamic expansion along the temporal dimension must be built upon preserving the input image's structure and ensuring its coherent evolution. The image defines the “starting point” of the visual narrative, while spatiotemporal consistency guarantees that the path from this point remains smooth and plausible.

        VideoCrafter1~\cite{chen2023videocrafter1opendiffusionmodels} is the first I2V-based model that can convert a given image into a video clip while maintaining the image content. It can strictly follow the content constraints of the provided reference image and preserve its content, structure and style. Ni et al.~\shortcite{ni2023conditional} proposed an I2V method of latent flow diffusion model (LFDM), which can synthesize optical flow sequences in the latent space according to given conditions to warp a given reference image. LFDM only needs to learn a low-dimensional latent space, which can improve computational efficiency. In order to enhance the guiding ability of image information, DynamiCrafter~\cite{xing2024dynamicrafter} uses a query converter to project it into a text-aligned rich context representation space, and connects the image information with the initial noise as input to enhance the consistency of the generated video. Framer~\cite{wang2024framer} and Wan2.1~\cite{wan2025wan} propose a method for video generation based on the first and last frames, aiming to produce video frames with smooth transitions between two images and achieve stronger video control capabilities.
        
        \paragraph{\bf Multi-scenes-guided Generation.}
        Multi-scenes-guided generation~\cite{melnik2024video,yang2023diffusion} aims to integrate complex inputs (such as narrative texts and multi-event prompts) to produce video content featuring dynamic scene transitions and overall visual consistency~\cite{chen2025skyreelsv2infinitelengthfilmgenerative}. This objective elevates task complexity to long-range spatiotemporal consistency. Multi-scene generation must satisfy stronger constraints across both spatial and temporal dimensions: spatially, it requires maintaining visual consistency within each independent scene; temporally, it must ensure seamless and natural transitions between different scenes. Consequently, multi-scene condition poses novel challenges to video generation: the ability to preserve both local spatiotemporal consistency and global narrative coherence.

        VideoDirectorGPT~\cite{lin2023videodirectorgpt} proposes to use LLM knowledge for video content planning and context-based video generation. It first performs scene planning through a video planner, and then controls the spatial layout through a generator, which can maintain temporal consistency in multiple scenes. TALC~\cite{bansal2024talc} further enhances the text adjustment mechanism in the T2V architecture to identify the temporal alignment between video scenes and scene descriptions. Each different video scene can be regarded as an event, and each event has its corresponding scene and entity. Wu et al.\shortcite{wu2025mind} proposed MinT, a multi-event video generator with time control. It binds each event to a specific time period in the generated video, and realizes time-aware interaction between event subtitles and video tags based on a special time position encoding method.

        \subsection{Multi-stage Generation Framework}
        Multi-stage generation framework~\cite{singer2022make,ge2023preserve} decomposes complex video generation tasks into a sequence of sequential or parallel sub-stages (such as base content creation, motion refinement, detail enhancement, etc). The advantage of this decomposition strategy lies in its ability to selectively address consistency constraints across spatial and temporal dimensions at different stages, thereby enabling hierarchical control and progressive optimization of spatiotemporal consistency in the final output.
        

        Cascade model~\cite{singer2022make,ge2023preserve,wang2025lavie} is a classic staged framework. In this model, the base module generates a sparse and low-resolution frame sequence. Then, a series of refinement modules (such as spatial and temporal super-resolution modules) are applied to enhance the resolution and frame rate of the output frames. Make-A-Video~\cite{singer2022make} and Imagen Video~\cite{ho2022imagen} design a spatiotemporal pipeline based on a cascaded video diffusion model, which includes a video encoder and decoder, an interpolation model, and a super-resolution model, capable of generating diverse videos and text animations in various artistic styles. Zhang et al.~\shortcite{zhang2023i2vgen} proposed I2VGen-XL, which separates video semantic accuracy from video quality to improve performance. Specifically, the base stage uses the CLIP visual encoder to ensure a high degree of semantic alignment. Then, the refinement stage enhances the details of the video through additional short text. VideoGen~\cite{li2023videogen} introduces an efficient I2V cascaded latent diffusion module that generates a latent video representation conditioned on a reference image and textual cues, followed by a flow-based temporal upsampling step to improve the temporal resolution.

        \subsection{Interactive Generation Framework}
        Interactive generation framework~\cite{lei2024comprehensive,yu2025surveyinteractivegenerativevideo} represents a user-centered paradigm that enables users to dynamically create personalized video content by intervening in generation models through real-time instructions. The core challenge lies in balancing immediate responsiveness to dynamic user input with maintaining the inherent spatiotemporal consistency of generated content. The interactive generation framework is particularly suitable for customized generation~\cite{skorokhodov2022stylegan}, such as interactive games~\cite{tang2025hunyuangamecraft2instructionfollowinginteractivegame}, embodied intelligence simulation, and autonomous driving.
        

        Genie 2~\cite{bruce2024genie} is a basic world model with video generation capabilities that can generate infinitely diverse and controllable 3D environments for training and evaluating embodied agents. It consists of an autoregressive dynamics model and a simple and scalable latent action model, allowing users to create frame by frame in the generated environment. In the field of games, GameGen-X~\cite{che2024gamegen} introduces multimodal control, which enables rich and flexible game content generation through text-to-character, sketch-to-environment, and brings a new gaming experience. GAIA-1~\cite{hu2023gaia} is an autonomous driving world model that combines video, text, and action inputs to generate realistic driving scenarios. It can not only generate high-quality videos, but also accurately control the generated scenes through text descriptions and action instructions, better meeting the testing needs of autonomous driving systems in different scenarios.

        \subsection{Summary}
        In summary, the diverse frameworks for video generation offer varied paradigms for sampling high-quality, highly consistent sequences from complex spatiotemporal distributions. Although these frameworks differ in their implementation mechanisms and focal points, their core objective remains the design of more effective sampling strategies and constraint forms. This enables the generated video sequences that meet requirements for visual realism, motion coherence, and semantic controllability.

\section{Post-processing Techniques}
\label{Sec:Post-processing Techniques}
         Following sequence sampling from high-dimensional spatiotemporal distributions to obtain initial video sequences, post-processing techniques~\cite{sun2024sora, wang2023video, dong2023video} may be regarded as a process of refining the distribution and optimizing the sequence. Due to constraints such as model capacity or sampling quantity, initial video sequences may exhibit local jitter, inconsistencies in detail, or temporal discontinuities. The core function of post-processing techniques lies precisely in performing local corrections and global smoothing on this set of sampled sequences to enhance spatiotemporal consistency.
         
        \subsection{Frame Interpolation}
            Frame interpolation technology~\cite{ge2023preserve,singer2022make, dong2023video, kye2025acevficomprehensivesurveyadvances} aims to enhance video frame rates and temporal continuity by synthesizing intermediate frames between adjacent frames, thereby mitigating motion judder and discontinuities commonly observed in generated videos.

            The principle of optical flow methods involves predicting pixel positions in subsequent frames by analyzing pixel changes between adjacent frames. Deep learning-based frame interpolation typically employs CNNs or GANs~\cite{ho2022imagen,ge2023preserve} to extract frame features and generate smooth intermediate frames. FlowNet~\cite{lipmanflow, ho2019flow++} pioneered the integration of CNNS with optical flow estimation, marking the first attempt to directly predict optical flow using networks. Xue et al.~\shortcite{xue2019video} introduced task-oriented flow, a motion representation learned through joint training that adapts to specific tasks such as frame interpolation and super-resolution. Niklaus et al.~\shortcite{niklaus2018context} proposed CtxSyn synthesizes intermediate frames by distorting neighboring frames and their pixel-level information through contextual mapping. The IFNet~\cite{huang2022real} architecture proposed by RIFE supports frame interpolation at arbitrary time steps via temporally encoded inputs. These methods enhance temporal smoothness and overall consistency in video sequences by preserving the spatiotemporal coherence of visual content during the generation of intermediate frames through estimating and modeling inter-frame motion dynamics.

        \subsection{Video Super-Resolution}
            Within the video generation, spatiotemporal super-resolution technology\cite{ho2022imagen} undertakes the critical task of reconstructing high-resolution video from low-resolution sequences. This technique aims to simultaneously enhance the fine spatial details of video frames while preserving the coherence as they evolve over the temporal dimension, thereby directly improving the spatiotemporal consistency of the generated video~\cite{ho2022imagen,ge2023preserve}.

            Diffusion models demonstrate potential in image super-resolution, yet their extension to video super-resolution tasks presents formidable challenges: models must not only perform appearance mapping from low to high resolution but also maintain temporal consistency between frames.
            MIA-VSR~\cite{zhou2024video} addresses this by designing intra- and inter-frame attention blocks that leverage augmented neighboring frame features to provide supplementary information. This approach enhances feature-level temporal continuity while reducing redundant computations. Upscale-A-Video~\cite{zhou2024upscale} constructs an implicit diffusion framework that preserves local consistency by integrating temporal layers within the U-Net and VAE decoder. It further enhances overall video stability through a training-free, optical flow-guided recurrent module that transmits and fuses implicit features globally. SATeCo~\cite{chen2024learning} proposes learning spatiotemporal guidance information from low-resolution video to calibrate latent-space denoising and pixel-space reconstruction processes, achieving precise temporal feature alignment.

        \subsection{Video Stabilization} 
            Video jitter represents a classic temporal inconsistency issue, manifesting as abrupt displacements, rotations, or deformations between frames. Such phenomena disrupt the smooth transition between adjacent spatiotemporal units within a sequence, severely compromising the video's spatiotemporal consistency. Video stabilization techniques~\cite{tulyakov2018mocogan,skorokhodov2022stylegan, wang2023video} achieve greater continuity and stability in the temporal dimension of sampled sequences by estimating and compensating for non-stationary motion between frames.

            SteadyFlow~\cite{yu2020learning} proposes a specific optical flow field with strong spatiotemporal consistency. By aggregating motion vectors along stable trajectories, it avoids fragile tracking of isolated feature points and can model spatially varying motion. Liu et al.~\shortcite{liu2016meshflow} focus on the problem of full-frame stabilization, proposing to synthesize stable frames by estimating dense deformation fields from adjacent frames and employing a learning-based hybrid spatial fusion strategy. This approach mitigates artifacts caused by inaccurate optical flow estimation or high-speed moving objects, enhancing the consistency of the generated videos. Shi et al.~\shortcite{shi2022deep} construct a joint motion representation by fusing sensor data with image optical flow, utilize LSTM to infer virtual camera poses, and ultimately achieve video stabilization through a deformable mesh, thereby improving the visual consistency of the video. Yang et al.~\shortcite{yang2023rerender} combined various advanced techniques to improve video rendering and enhancement functions. The work employed motion estimation and compensation algorithms to identify and eliminate jitter in videos, thereby improving stability. 

        \subsection{Video Deblurring}
             Video blurring~\cite{liu2024sora} is typically caused by camera or object motion, manifesting in the spatiotemporal domain as the mixing and loss of information. If deblurring is applied to individual frames in isolation, inconsistencies in detail may arise due to the neglect of inter-frame relationships, thereby compromising temporal continuity. Video deblurring aims to recover clear content from sequences of blurred frames while preserving the spatiotemporal consistency of the spatiotemporal unit sequence. It is widely utilized in high-resolution imaging applications~\cite{he2025domain}, such as medical imaging, film generation, and other fields, to improve the viewing experience.

             Zhou et al.~\shortcite{zhou2019spatio} proposed the Spatiotemporal Filtering Adaptive Network (STFAN), which extracts blur features from previously blurred and restored images and aligns them with the current frame to eliminate blur caused by feature space variations. Lin et al.~\shortcite{lin2022flow} achieved efficient modeling of spatiotemporal features by combining motion estimation with attention mechanisms. Simultaneously, they introduced a recurrent embedding mechanism to establish long-range temporal dependencies through inter-frame hidden state propagation, thereby expanding the temporal receptive field. Rao et al.~\shortcite{rao2024rethinking} proposed VD-Diff, integrating diffusion models into a wavelet-aware dynamic transformer. This approach preserves and restores low-frequency information in videos while simultaneously generating high-frequency detail. He et al.~\shortcite{he2025domain} proposed a domain-adaptive deblurring model scheme. By extracting relatively clear regions as pseudo-clear images and combining them with deblurring techniques to generate blurred images, they achieved domain adaptation for unseen domains.

        \subsection{Summary}
         In summary, post-processing techniques systematically enhance the spatiotemporal consistency of video by applying local corrections and global smoothing to the initial sampling sequence. This constitutes a crucial optimization step in the progression from ‘initial generation’ to ‘high-quality output’. Technologically, it compensates for the limitations of generation models in dynamic modeling and long-term coherence, effectively ensuring the spatiotemporal consistency of video generation.

\section{Training Strategies}
\label{Sec:Training Strategies}
     The training strategy~\cite{melnik2024video,xiong2024autoregressive,liu2024sora} of video generation models aims to optimize the model's performance in terms of spatiotemporal consistency, generation quality, and computational efficiency, while adapting to a variety of application scenarios. An effective training strategy~\cite{sun2024sora} can improve the model's ability to model complex video content and enhance its ability to generate coherent video sequences. This section reviews a variety of common training strategies, including Transfer Learning~\cite{wang2023modelscope}, Progressive Learning~\cite{hong2022cogvideo}, Image and Video Joint Learning~\cite{polyak2024movie}, Local Module Adaptation~\cite{wang2025easycontrol}, Training-free Learning~\cite{qi2023fatezero}, Model Distillation~\cite{ho2022imagen} and Reward Feedback Learning~\cite{yuan2024instructvideo}. These strategies address the challenges in video generation through different optimization methods and data utilization mechanisms.
    \subsection{Transfer Learning}
    In the video generation task, transfer learning~\cite{melnik2024video} uses a model pre-trained on image datasets to provide initialized parameters for the video generation model, which contains rich visual and dynamic features. This enables the model to quickly adapt to complex tasks to accelerate the training process and improve model performance.

     In Stable Video Diffusion~\cite{blattmann2023stablevideodiffusionscaling} work, it proposes three stages for training video latent diffusion models: text-to-image pre-training, video pre-training, and high-quality video fine-tuning. Text-to-image pre-training enables the model to master the ability to generate static images and improve the model's initial feature extraction ability. The next two stages are fine-tuned based on video datasets to achieve the transfer of model capabilities. Among them, in the video pre-training stage, the model can learn the temporal dynamics and motion patterns between video frames. In the high-quality video fine-tuning stage, the model is fine-tuned on high-resolution or domain-specific datasets to further optimize the generation quality and detail performance. ModelScopeT2V~\cite{wang2023modelscope} and Sora~\cite{openai2024sora} both demonstrate the effectiveness of transfer learning. First, a visual feature extractor is constructed using text-to-image pre-training, and then the temporal module is introduced and fine-tuned on a high-quality video dataset to optimize the generation ability of complex scenes. In the SimDA~\cite{xing2024simda} work, a lightweight spatiotemporal adapter is designed for transfer learning. In order to improve the temporal consistency of video generation, a latent shift attention mechanism (LSA) is proposed to improve the temporal perception ability.


    \subsection{Progressive Learning}
    Progressive learning~\cite{yang2023diffusion,xing2024survey} improves training stability by gradually increasing training complexity. The method starts with simpler tasks, such as short clips, which require less computing resources and are easier to learn. As training progresses, the model is gradually exposed to more complex tasks, such as longer videos, or more complex motion patterns and texture details~\cite{wang2024loong}. This step-by-step approach allows the model to first grasp the basic characteristics of the video than tackling higher-order complexities.

    CogVideoX~\cite{hong2022cogvideo} adopts a progressive training strategy from simple to complex, focusing on low-resolution or short-duration videos in the early stage, and gradually increasing the resolution and video duration. At the same time, it combines dynamic frame packaging to make it adaptable to different video resolutions. Loong~\cite{wang2024loong} takes into account the loss function imbalance caused by the increase in the length of the video sequence, and introduces a loss function reweighting mechanism to ensure that the model is evenly optimized over the entire video sequence by dynamically adjusting the loss weights of different time steps.
    
    \subsection{Image and Video Joint Learning}
    Considering that available video data is scarce and of low quality, high-quality image data is relatively abundant~\cite{singh2023survey}. Image and video joint learning makes up for this limitation by combining the two, enabling the model to achieve better performance on limited video data. Static images enable the model to adapt to different scenes and content, while video data enhances the model's ability to understand dynamic scenes~\cite{liu2024sora}.


    Image and video data have inherently different characteristics. The goal of joint training is to unify the data of these two modalities through a shared latent space, so that the model can learn both static and dynamic features~\cite{lin2024video}. Image-Video Joint VAE is a key tool that encodes images and videos in RGB pixel space into a spatiotemporally compressed latent representation through a VAE framework and performs generation tasks in this latent space. Movie Gen Video~\cite{polyak2024movie} and Goku~\cite{chen2025gokuflowbasedvideo} adopt a spatiotemporal compression strategy, which compresses the width, height, and time dimensions of the video separately through downsampling, significantly reducing the dimensionality of the latent representation, thereby reducing training costs and improving training and inference efficiency.

    
    \subsection{Local Module Adaptation}
    Local module adaptation~\cite{singh2023survey,zhou2024survey} fine-tunes specific components in the model (such as motion modules and spatial modules) to better adapt them to specific target video domains. This method significantly improves the performance of the model in specific tasks or domains while maintaining the general knowledge of the pre-trained model.


    In the AnimateDiff~\cite{guo2024animatediff} work, a plug-and-play motion module is used, which can be seamlessly integrated into any personalized T2V model after only one training. In addition, based on the lightweight fine-tuning technology MotionLoRA, the motion module can adapt to new motion patterns with low training and data collection costs. EasyControl~\cite{wang2025easycontrol} introduces a conditional adapter to extract conditional features and inject them into the interchangeable basic text-to-video model to guide video generation.

    
    \subsection{Training-Free Learning}
    Training-free learning~\cite{melnik2024video,qiufreenoise,lu2024freelong} is a video generation strategy that does not require specialized training for a specific task. It relies on the strong generalization ability of pre-trained models and combines external guidance signals (such as text prompts, visual embeddings, etc.) to achieve high-quality video generation.

    FateZero~\cite{qi2023fatezero} is a zero-shot text editing method that can be used for real-world video editing without per-prompt training. Zhang et al.~\shortcite{zhang2025videomerge} proposed VideoMerge, a training-free method that can be seamlessly adapted to merge with pre-trained video models. The method retains the original expressiveness and consistency of the model while allowing for extended duration and dynamic changes. MOVi~\cite{rahman2025movi} adopts a training-free method for multi-object video generation, which uses LLM to guide the motion trajectory of objects to achieve precise control of real motion and better captures object-specific features and motion patterns through a cross-attention mechanism.
    
    \subsection{Model Distillation}
    Model distillation~\cite{shao2025magicdistillation} achieves performance close to that of a teacher model by having a smaller student model learn from a larger teacher model. The teacher model is a model with large parameters and high computational complexity. It has strong expressive power and high performance, but slow reasoning speed and high resource consumption. The student model is designed to be lighter and suitable for deployment in resource-constrained scenarios~\cite{li2024survey}. It can achieve similar video generation quality at a lower computational cost.


    Imagen Video~\cite{ho2022imagen} uses progressive distillation technology to improve the efficiency of video reasoning generation. Progressive distillation achieves fast and high-quality video sampling without classifier guidance by reducing sampling steps in multiple stages. Magic 1-For-1~\cite{yi2025magic} decomposes the complex text-to-video generation task into two independent subtasks (text-to-image, image-to-video). Distillation technology is applied to each subtask to reduce the sampling steps of the diffusion model. By training the student model to imitate the output distribution of the teacher model, Magic 1-For-1 is able to complete high-quality generation in very few sampling steps. MagicDistillation~\cite{shao2025magicdistillation} focuses on optimizing large video diffusion models through weak-to-strong distillation technology to achieve efficient video synthesis. It specifically targets portrait video generation and improves generation quality through distribution matching distillation (DMD) and LoRA.

    
    \subsection{Reward Feedback Learning}
    Reward feedback learning~\cite{sun2024sora} is a machine learning optimization method that aims to guide the model to generate outputs that meet specific goals through external feedback signals (from humans or automated systems). In the field of video generation, reward feedback learning defines a reward function to quantify the quality of the generated video, and uses reinforcement learning or other optimization techniques to adjust the model parameters so that the generated results are more in line with user needs or task goals~\cite{xu2024visionreward}.

    InstructVideo~\cite{yuan2024instructvideo} guides the text-to-video diffusion model with human feedback through reward fine-tuning. It mitigates the performance degradation of temporal modeling and improves the efficiency of fine-tuning by designing segmented video rewards and time-decayed rewards. Control-A-Video~\cite{chen2024controlavideocontrollabletexttovideodiffusion} proposes a spatiotemporal reward feedback learning (ST-ReFL) algorithm that uses multiple reward models to optimize the video diffusion model to improve video quality and motion consistency. Huang et al.~\shortcite{huang2024diffusion} express diffusion reward as the negative value of conditional entropy to encourage effective exploration of expert behavior. It can be effectively applied to manipulation tasks of robots in simulation platforms and real world. VisionReward~\cite{xu2024visionreward} realizes the alignment of visual generation models with human preferences. It decomposes human preferences in a fine-grained manner and scores them in the form of judgment questions, and finally weights and sums the scores to achieve effective preference prediction.

    \subsection{Summary}
    In summary, training strategies for video generation models focus on optimizing both output quality and computational efficiency while specifically addressing the core challenge of spatiotemporal consistency. These approaches not only enhance inter-frame content coherence and dynamic naturalness but also strengthen structural stability during long-sequence generation. Looking ahead, with further advancements in training paradigms such as causal temporal learning and physical constraint injection, video generation models are poised to achieve highly consistent spatiotemporal capabilities in complex real-world scenarios.

    \begin{table*}[]

\centering
\scalebox{0.8}{
\begin{tabular}{ccccccc}
\hline
Class                                      & Benchmark         & Year & T2V & I2V & Open-Domain & Group                             \\ \hline
\multirow{2}{*}{Text alignment}            & StoryBench~\cite{bugliarello2023storybench}        & 2023 & \color{green}{\ding{52}}   & \color{red}{\ding{56}}   & \color{green}{\ding{52}}           & Google Research                   \\ \cline{2-7} 
                                           & StoryEval~\cite{wang2025your}         & 2025 & \color{green}{\ding{52}}   & \color{red}{\ding{56}}   & \color{green}{\ding{52}}           & University of Washington          \\ \hline
\multirow{6}{*}{Temporal dynamics}         & MiraBench~\cite{ju2024miradata}         & 2024 & \color{green}{\ding{52}}   & \color{red}{\ding{56}}   & \color{green}{\ding{52}}           & Tencent                           \\ \cline{2-7} 
                                           & DEVIL~\cite{liao2024evaluation}             & 2024 & \color{green}{\ding{52}}   & \color{red}{\ding{56}}   & \color{green}{\ding{52}}           & Dalian University of Technology   \\ \cline{2-7} 
                                           & T2VBench~\cite{ji2024t2vbench}          & 2024 & \color{green}{\ding{52}}   & \color{red}{\ding{56}}   & \color{green}{\ding{52}}           & Carnegie Mellon University        \\ \cline{2-7} 
                                           & ChronoMagic-Bench~\cite{yuan2024chronomagic} & 2024 & \color{green}{\ding{52}}   & \color{red}{\ding{56}}   & \color{green}{\ding{52}}           & University of California          \\ \cline{2-7} 
                                           & TC-Bench~\cite{feng2024tc}          & 2024 & \color{green}{\ding{52}}   & \color{green}{\ding{52}}   & \color{green}{\ding{52}}           & 1University of California         \\ \cline{2-7} 
                                           & VMBench~\cite{ling2025vmbench}           & 2025 & \color{green}{\ding{52}}   & \color{red}{\ding{56}}   & \color{green}{\ding{52}}           & Alibaba                           \\ \hline
\multirow{7}{*}{Comprehensive performance} & LFDM Eval~\cite{ni2023conditional}         & 2023 & \color{green}{\ding{52}}   & \color{red}{\ding{56}}   & \color{red}{\ding{56}}           & The Pennsylvania State University \\ \cline{2-7} 
                                           & CATER-GEN~\cite{hu2022make}         & 2023 & \color{green}{\ding{52}}   & \color{green}{\ding{52}}   & \color{red}{\ding{56}}           & Wuhan University                  \\ \cline{2-7} 
                                           & EvalCrafter~\cite{liu2024evalcrafter}       & 2023 & \color{green}{\ding{52}}   & \color{green}{\ding{52}}   & \color{green}{\ding{52}}           & Tencent                           \\ \cline{2-7} 
                                           & AIGCBench~\cite{fan2023aigcbench}         & 2024 & \color{green}{\ding{52}}   & \color{green}{\ding{52}}   & \color{green}{\ding{52}}           & Chinese Academy of Sciences       \\ \cline{2-7} 
                                           & Video-Bench~\cite{han2025video}       & 2025 & \color{green}{\ding{52}}   & \color{green}{\ding{52}}   & \color{green}{\ding{52}}           & Shanghai Jiao Tong University     \\ \cline{2-7} 
                                           & VBench~\cite{huang2024vbench}            & 2025 & \color{green}{\ding{52}}   & \color{green}{\ding{52}}   & \color{green}{\ding{52}}           & Nanyang Technological University  \\ \cline{2-7} 
                                           & T2V-CompBench~\cite{sun2025t2v}     & 2025 & \color{green}{\ding{52}}   & \color{red}{\ding{56}}   & \color{green}{\ding{52}}           & The University of Hong Kong       \\ \hline
\end{tabular}
}
\caption{Sumary of Benchmarks}
\label{tab:Sumary of Benchmarks}
\vspace{-3em}
\end{table*}

\section{Benchmarks and Evaluation Metrics}
\label{Sec:Benchmarks and Evaluation Metrics}
    This section will explore the evaluation of spatiotemporal modeling capabilities of video generation models, which is an important research topic that aims to measure the performance of models in capturing temporal dynamics and spatial information in video sequences~\cite{liu2024sora,sun2024sora}. Specifically, we will focus on the benchmarks and evaluation metrics in this field to systematically compare and analyze the performance of different models.

    \subsection{Benchmarks}
    Benchmark is a standardized test framework for evaluating video generation models~\cite{fan2023aigcbench,ji2024t2vbench}. It carefully designs datasets and tasks to evaluate the ability of video generation in simulated scenarios. We summarize some popular benchmark works, as shown in Table \ref{tab:Sumary of Benchmarks}.


    \paragraph{\bf Text Alignment:}
    Benchmarks focused on text alignment evaluate the model's ability to generate videos that accurately correspond to text descriptions. Among them, story-oriented benchmarks~\cite{sun2024sora} are often used to evaluate the story generation and text alignment capabilities of T2V models. StoryBench~\cite{bugliarello2023storybench} and StoryEval~\cite{wang2025your} are specifically designed to evaluate the model's ability to handle story-level video generation tasks. They parse text prompts based on advanced visual language models and establish an evaluation framework for evaluating video stories, which can evaluate the model's ability to migrate from simple instructions to complex story construction.
    
    \paragraph{\bf Temporal Dynamics:}
    Benchmarks under temporal dynamics focus on evaluating a model's ability to capture and generate coherent temporal sequences, ensuring smooth transitions and realistic motion in videos. ChronoMagic-Bench~\cite{yuan2024chronomagic} and T2VBench~\cite{ji2024t2vbench} focus on the temporal coherence and rationality of temporal logic of generated videos, and use automated evaluation metric to analyze the temporal coherence of generated videos. MiraBench~\cite{ju2024miradata} and DEVIL~\cite{liao2024evaluation} build high-quality video datasets to analyze the naturalness of motion of videos generated by the T2V model and verify whether it conforms to human intuition.

    \paragraph{\bf Comprehensive Performance:}
    Benchmarks for comprehensive performance aim to evaluate video generation models holistically, combining aspects of text alignment, temporal dynamics and visual quality~\cite{liu2024sora,li2024survey}. EvalCrafter~\cite{liu2024evalcrafter} is a framework for evaluating T2V generation. It evaluates the most advanced video generation models in terms of visual quality, content quality, motion quality, and text alignment based on 17 objective indicators. AIGCBench~\cite{fan2023aigcbench} focuses on I2V generation. It builds a diverse dataset based on GPT-4 and T2I models, and then evaluates the generation model based on four dimensions. Video-Bench~\cite{han2025video} provides a structured and scalable video evaluation strategy by combining small-shot scoring and query chaining techniques to achieve consistent alignment with human preferences. VBench~\cite{huang2024vbench} is a comprehensive video generation benchmark suite that divides video evaluation into fine-grained categories and constructs fine-level evaluation indicators, studying the capabilities of current models in various dimensions. For synthetic text, Sun et al.~\shortcite{sun2025t2v} proposed T2V-CompBench, which designs relevant evaluation indicators based on different aspects of compositionality, which can better reflect the quality of synthetic text-to-video generation.


    \subsection{Evaluation Metrics}
    This section summarizes some evaluation metrics for video generation and divides them into three categories: frame quality assessment~\cite{li2024survey}, video smoothness assessment~\cite{liu2024sora}, and overall video assessment~\cite{jiangkuo2024chinese}. Among these metrics, frame quality assessment focuses on the consistency of static spatial elements within individual frames. Video smoothness assessment assesses temporal consistency and the rationality of motion across the video. Overall video assessment examines the coherence and quality of the entire frame sequence as a unified whole. 

   
    \paragraph{\bf Frame Quality Assessment} 
    Frame quality evaluation~\cite{cho2024sora} is performed on a video frame basis, evaluating the generation quality at both the pixel level and semantic level of the generated frame. here are some commonly used frame quality assessment metrics:
    \begin{itemize}

    \item{\bf PSNR (Peak Signal-to-Noise Ratio):} PSNR~\cite{he2025domain} is a widely adopted metric for evaluating the quality of generated images. It quantifies the ratio between the maximum possible pixel value of the original image and the mean squared error (MSE) between the pixel values of the original and generated images. Expressed in decibels (dB), a higher PSNR value indicates better image fidelity, reflecting smaller differences between the two images.

    \item{\bf SSIM (Structural Similarity Index):}  SSIM~\cite{li2024survey} evaluates the similarity between original and generated images by considering changes in luminance, contrast, and structural information. It assesses perceptual quality by modeling human visual perception, capturing how well the structural features of the generated image align with the original. Higher SSIM values indicate greater similarity.

    \item{\bf IS (Inception Score):} IS~\cite{bar2024lumiere} leverages a pre-trained Inception V3 neural network to evaluate two key aspects of generated images: quality and diversity. Among them, the quality evaluation is based on the Inception V3 classifier to evaluate whether the generated image can make reliable predictions for specific categories to measure its recognizability. And the diversity evaluation is by analyzing the distribution of predicted category probabilities in a set of generated samples.

     \item{\bf Fréchet Inception Distance (FID):} FID~\cite{blattmann2023align} uses a network to extract abstract features of images and analyzes the distance between the generated and original images in feature space, reflecting the distance between the generated distribution and the original distribution. The FID value reflects the difference between the two distributions. The smaller the value, the higher the similarity between the generated distribution and the original distribution.

     \item{\bf Aesthetic Score:} The aesthetic score~\cite{huang2024vbench} is a critical metric for evaluating the visual appeal of generated images or video frames, focusing on subjective qualities like composition, color harmony, and overall pleasantness. It uses a pre-trained aesthetic detector to score the generated images or videos. The higher the score, the more the generated result conforms to human aesthetics.

    \end{itemize}

    \paragraph{\bf Video Smoothness Assessment} 
    Video smoothness assessment evaluates the temporal coherence and naturalness of motion in a video sequence, ensuring that the generated video exhibits seamless frame-to-frame transitions and plausible dynamic behavior.
    \begin{itemize}
    \item {\bf Optical Flow Consistency:} Optical flow consistency~\cite{teed2020raft} is based on the optical flow estimation model (such as RAFT) to predict the optical flow, and hopes to minimize the distance between the predicted optical flow value and the true value. This indicator can effectively measure the smoothness of the generated video.

    \item{\bf Dynamic Coherence:} Dynamic consistency~\cite{huang2024vbench} is used to evaluate the consistency of the action evolution of the video over time with the given input conditions (such as text prompts describing a specific action sequence), and it is expected that the video actions can be performed strictly according to the prompts.

    \item{\bf Motion Rationality:} Motion rationality~\cite{ling2025vmbench} evaluates whether the motion in a generated video adheres to logical and physically realistic principles, ensuring that actions and dynamics are coherent and contextually appropriate. This evaluation can be completed using a multi-question answering pipeline based on video content, by prompting GPT to generate a visual description of the expected physical behavior or motion to judge the rationality. 

    \item{\bf Temporal Flickering:} In video generation, there are often inconsistencies or jitters between frames, resulting in visual incoherence. Temporal flickering~\cite{huang2024vbench} reflects the severity of the flicker phenomenon by calculating the local difference between two adjacent frames. The smaller the value, the less flicker in the generated video and the better the temporal smoothness.
    \end{itemize}

    \paragraph{\bf Overall Video Assessment}
   Overall video assessment~\cite{zhou2024survey} is a comprehensive evaluation method that takes the entire video or a sampled partial video as the evaluation unit and considers the overall spatiotemporal and semantic features of the generated video to evaluate its quality. 
    \begin{itemize}
    \item {\bf Basic Video Quality:} Focusing on the basic visual issues of generated videos, such as picture distortion, noise or artifacts. VAMF~\cite{Orduna2020VMAF} is a commonly used measurement metric. It uses machine learning algorithms to merge basic evaluation metrics into a final metrics, and adapts weights for each metric to finally obtain a comprehensive quality score. 

     \item{\bf Content Consistency:} Generated videos generally need to be consistent in terms of subject, background, human ID, etc~\cite{huang2024vbench}. Generally, a specific pre-trained model is used to extract video frame information and determine the similarity between frames to determine whether consistency can be maintained.

    \item{\bf Feature Similarity:} Drawing on FID in the image field, FVD~\cite{ge2024content} is proposed for video evaluation, which avoids the shortcomings of frame-by-frame indicators by estimating the feature distribution distance between the generated video and the real one. From the results, the smaller the FVD, the higher the similarity of video features.

    \item{\bf Conditional Alignment:} The generated video needs to be aligned with the provided text prompts and reference images~\cite{ji2024t2vbench}. Generally, the video frames are sampled first, and then the average similarity with the conditional feature vector is calculated as the alignment metric. 

    \item{\bf Time Sequence Similarity:} Considering the video as a time sequence, this metric measures the similarity between the generated sequence and the real sequence~\cite{lahreche2021comparison}. Commonly, it is divided into equal length cases and unequal length cases. For equal length sequences, the Euclidean Distance and Pattern Distance of the corresponding elements are usually calculated. For unequal length sequences, the DTW algorithm~\cite{bringmann2024dynamic} realizes sequence alignment and calculates its sequence distance based on the idea of dynamic programming.
    
    \end{itemize}

    \subsection{Summary}
    In summary, by establishing a systematic benchmark and evaluation metric system, we can objectively quantify a model's performance across spatiotemporal consistency, motion naturalness, and structural fidelity. Moving forward, as more discriminative assessment methods that closely align with human perception emerge, this will become a crucial foundation for advancing the field and ensuring the reliability and usability of generated content.

\section{Future Directions and Challenges}
\label{Sec:Future Directions and Challenges}
        In recent years, as user demands have grown increasingly sophisticated, video generation has advanced rapidly towards greater complexity and diversity. This evolution presents both opportunities and challenges. These developments have elevated spatiotemporal consistency, which is a core challenge, from a technical hurdle to a critical bottleneck for achieving genuine visual realism and narrative coherence. The difficulties manifest across various cutting-edge domains as follows: 
        \paragraph{\bf Long Video Generation.} Long video generation~\cite{li2024survey} represents a significant developmental trend for content applications such as short films, game animations, and online education. However, as sequence lengths increase, the generation task extends beyond maintaining frame-by-frame continuity to achieving long-range spatiotemporal consistency. For instance, during generation spanning thousands of frames or even across scenes, it is essential to consistently preserve the unity of multi-dimensional information including character identity, object attributes, and environmental states. Existing generation models typically struggle to effectively capture such long-range dependencies, lacking the dynamic modeling and spatiotemporal memory capabilities required for complex and extended-duration relationships. Concurrently, long video entails modeling extremely high-dimensional spatiotemporal joint distributions, presenting significant challenges in computational memory, training stability, and inference efficiency. Furthermore, training generation models with strong generalization capabilities under scarce large-scale data and prohibitively high annotation costs remains a major obstacle to the practical applications. Effectively extracting and preserving spatiotemporal causal and semantic associations within lengthy sequences remains a core challenge requiring urgent breakthroughs.
        \paragraph{\bf Personalized Video Generation.} Personalized generation~\cite{ye2024stylemaster} aims to tailor videos based on specific user inputs. Its reliance on multi-modal signals—such as text, images, and sketches—exacerbates the difficulty of cross-modal semantic alignment. This often leads to control signals being weakened or distorted during long-term generation, thereby undermining spatiotemporal consistency. Concurrently, systems encounter dual constraints when responding to dynamic user instructions: on one hand, responses to fine-grained adjustments (e.g., specific actions, facial expressions) frequently produce localized distortions or semantic deviations; on the other, during extensive modifications (e.g., overall background changes, style transitions), models struggle to effectively coordinate new and existing elements while preserving the spatiotemporal consistency of generated content, often resulting in residual inertia or logical discontinuities. This highlights how existing personalized generation methods struggle to balance precise control with spatiotemporal consistency, particularly in preserving long-range information and coordinating dynamic content.
        \paragraph{\bf Video Emotion Expression.} The objective of video generation extends beyond merely depicting scenes; it aims to evoke emotions and narrate compelling narratives~\cite{liu2024sora}. Emotional expression is deeply intertwined with expressive spatiotemporal consistency. This demands that foundational consistency, such as stable objects or fluid motion, serves higher-level emotional consistency. For instance, a tense atmosphere must be sustained through consistent lighting, color tones, and editing rhythms across shots; a character's emotional state must be presented temporally coherently through facial expressions, body language, and interactions. Inconsistent emotional cues, such as abrupt shifts in musical tone or unexplained character mood swings, fail to resonate with audiences. Therefore, the model must learn to generate narrative sequences that are temporally and spatially coherent and semantically and emotionally consistent, ensuring that the underlying pixel dynamics align with the higher-level emotional content.
        \paragraph{\bf Video World Model.} The core of such world models~\cite{sora,liu2024sora} lies in internalizing spatiotemporal consistency as an understanding of worldly principles. By learning from vast video datasets, they construct an internal representation that implicitly comprehends three-dimensional geometry, object persistence, and the causality of motion. Generating or predicting based on this “world knowledge” fundamentally ensures the plausibility of long-range motion and the logical coherence of scene transitions~\cite{baldassarre2025featuresdinofoundationvideo}. Current primary challenges lie in: firstly, striking a balance between spatiotemporal  feature representation and detail fidelity; secondly, the still-limited capacity to simulate complex world interactions and long-range causality. The ultimate goal is to become a powerful simulator capable of generating virtual realities that genuinely conform to the laws of the world.
        \paragraph{\bf Video Generation Evaluation.} The lack of adequate evaluation metrics severely hampers progress in addressing the aforementioned consistency challenges~\cite{ji2024t2vbench, huang2024vbench}. Current metrics are largely migrated from image generation domains, focusing on single-frame quality or short-term paired similarity, and failing to capture the essence of video as a temporal whole. The field urgently requires benchmarks capable of directly quantifying spatiotemporal inconsistencies. This encompasses metrics for long-range identity preservation, motion smoothness and realism, narrative coherence, and temporal style stability. Developing such a comprehensive evaluation framework requires not only novel computational metrics but also large-scale human perception studies to assess consistency. Without these tools, it is difficult to diagnose model failures, compare technological advancements, or drive research specifically addressing the profound consistency challenges in long-form, personalized, and emotional video generation.







\section{Conclusions}
\label{Sec:Conclusions}
        Spatial and temporal consistency is one of the key challenges in the video generation process. In this survey, we provide a comprehensive summary of recent video generation techniques from the perspective of spatiotemporal consistency and analyzes their contributions to maintaining this consistency. We conclude them in the following: 1) efficient spatiotemporal feature representation is beneficial for complex video generation tasks; 2) The design of video generation framework and training strategy needs to take into account both spatial consistency and temporal consistency; 3) video generation is gradually evolving toward more complex and refined directions, which will emerge both new opportunities and challenges. We hope that our work will make a meaningful contribution to the future of video generation.
        \par


\bibliographystyle{ACM-Reference-Format}
\bibliography{Reference}


\appendix

\section{Summary of Related Works}
The reference work summary for this paper is shown in Table \ref{tab:Summary of Video Works}.
\begin{table*}[]
\centering
\scalebox{0.54}{ 

\begin{tabular}{@{}cccccccccc@{}}
\toprule
\multirow{2}{*}{Model Name} & \multirow{2}{*}{Year} & \multirow{2}{*}{Param}                          & \multirow{2}{*}{Representation} & \multirow{2}{*}{Discretization} & \multirow{2}{*}{Backbone} & \multirow{2}{*}{Training Strategy}                                                            & \multicolumn{2}{c}{Support} & \multirow{2}{*}{Group}                                                                        \\ \cmidrule(lr){8-9}
                            &                       &                                                 &                                 &                                 &                           &                                                                                               & T2V          & I2V          &                                                                                               \\ \midrule
VideoGPT~\cite{yan2021videogpt}                    & 2021                  & -                                               & VQ-VAE                          & \color{green}{\ding{52}}                               & Autoregressive            & Transfer learning                                                                             & \color{red}{\ding{56}}            & \color{green}{\ding{52}}            & UC Berkeley                                                                                   \\ \midrule
Make-A-Video~\cite{singer2022make}                & 2022                  & 9.6B                                            & BPE, CLIP                       & \color{red}{\ding{56}}                               & Diffusion                 & Transfer learning                                                                             & \color{green}{\ding{52}}            & \color{red}{\ding{56}}            & Meta AI                                                                                       \\ \midrule
NUWA-Infinity~\cite{wu2022nuwab}               & 2022                  & -                                               & VQGAN                           & \color{red}{\ding{56}}                               & Autoregressive            & Image and video joint learning                                                                & \color{green}{\ding{52}}            & \color{red}{\ding{56}}            & Microsoft Research Asia                                                                       \\ \midrule
Imagen Video~\cite{ho2022imagen}                & 2022                  & 11.6B                                           & T5                              & \color{red}{\ding{56}}                               & Diffusion                 & \begin{tabular}[c]{@{}c@{}}Image and video joint learning\\ Model distillation\end{tabular}   & \color{green}{\ding{52}}            & \color{red}{\ding{56}}            & Google Research                                                                               \\ \midrule
Video LDM~\cite{blattmann2023align}                   & 2023                  & -                                               & 3D-CNN                          & \color{red}{\ding{56}}                               & Diffusion                 & Transfer learning                                                                             & \color{green}{\ding{52}}            & \color{red}{\ding{56}}            & LMU Munich                                                                                    \\ \midrule
LFDM~\cite{ni2023conditional}                        & 2023                  & -                                               & LFAE                            & \color{red}{\ding{56}}                               & Diffusion                 & Transfer learning                                                                             & \color{red}{\ding{56}}            & \color{green}{\ding{52}}            & The Pennsylvania State University                  \\ \midrule
ModelScope~\cite{wang2023modelscope}                  & 2023                  & 1.7B                                            & VQGAN                           & \color{green}{\ding{52}}                               & Diffusion                 & Direct learning                                                                               & \color{green}{\ding{52}}            & \color{red}{\ding{56}}            & Alibaba                                                                                       \\ \midrule
Stable Video Diffusion~\cite{blattmann2023stablevideodiffusionscaling}      & 2023                  & 1B                                              & VAE                             & \color{red}{\ding{56}}                               & Diffusion                 & Transfer learning                                                                             & \color{green}{\ding{52}}            & \color{red}{\ding{56}}            & Stability AI                                                                                  \\ \midrule
Nuwa-XL~\cite{yin2023nuwa}                     & 2023                  & -                                               & VAE                             & \color{red}{\ding{56}}                               & Diffusion                 & -                                                                                             & \color{green}{\ding{52}}            & \color{red}{\ding{56}}            & University of Science and Technology of China      \\ \midrule
MagicVideo~\cite{zhou2023magicvideoefficientvideogeneration}                  & 2023                  & -                                               & VAE                             & \color{red}{\ding{56}}                               & Diffusion                 & Direct learning                                                                               & \color{green}{\ding{52}}            & \color{red}{\ding{56}}            & ByteDance                                                                                     \\ \midrule
LaVie~\cite{wang2025lavie}                  & 2023                  & -                                               & VAE,CLIP                             & \color{red}{\ding{56}}                               & Diffusion                 & Direct learning                                                                               & \color{green}{\ding{52}}            & \color{red}{\ding{56}}            & Shanghai Artificial Intelligence Laboratory                                                                                     \\ \midrule
Tune-A-Video~\cite{wu2023tune}                & 2023                  & -                                               & VAE                             & \color{red}{\ding{56}}                               & Diffusion                 & Transfer learning                                                                             & \color{green}{\ding{52}}            & \color{red}{\ding{56}}            & Show Lab                                                                                      \\ \midrule
Rerender-A-Video~\cite{yang2023rerender}            & 2023                  & -                                               & VQ-VAE                          & \color{green}{\ding{52}}                               & Diffusion                 & Training-free learning                                                                        & \color{green}{\ding{52}}            & \color{red}{\ding{56}}            & S-Lab                                                                                         \\ \midrule
VideoCrafter1~\cite{chen2023videocrafter1opendiffusionmodels}               & 2023                  & -                                               & VAE                             & \color{red}{\ding{56}}                               & Diffusion                 & Transfer learning                                                                             & \color{green}{\ding{52}}            & \color{green}{\ding{52}}            & Tencent                                                                                       \\ \midrule
VideoDirectorGPT~\cite{lin2023videodirectorgpt}            & 2023                  & -                                               & CLIP                            & \color{red}{\ding{56}}                               & Diffusion                 & Transfer learning                                                                             & \color{green}{\ding{52}}            & \color{red}{\ding{56}}            & UNC Chapel Hill                                                                               \\ \midrule
I2VGen-XL~\cite{zhang2023i2vgen}                   & 2023                  & -                                               & VQGAN                           & \color{green}{\ding{52}}                               & Diffusion                 & Transfer learning                                                                             & \color{red}{\ding{56}}            & \color{green}{\ding{52}}            & Alibaba                                                                                       \\ \midrule
VideoGen~\cite{li2023videogen}                    & 2023                  & -                                               & CLIP                            & \color{red}{\ding{56}}                               & Diffusion                 & Local module adaptation                                                                       & \color{green}{\ding{52}}            & \color{red}{\ding{56}}            & Baidu                                                                                         \\ \midrule
GAIA-1~\cite{hu2023gaia}                      & 2023                  & 6.5B                                            & CLIP, T5                        & \color{green}{\ding{52}}                               & Diffusion                 & Image and video joint learning                                                                & \color{green}{\ding{52}}            & \color{green}{\ding{52}}            & Wayve                                                                                         \\ \midrule
Control-A-Video~\cite{chen2024controlavideocontrollabletexttovideodiffusion}             & 2023                  & -                                               & -                               & \color{red}{\ding{56}}                               & Diffusion                 & Reward feedback learning                                                                      & \color{green}{\ding{52}}            & \color{green}{\ding{52}}            & ByteDance                                                                                     \\ \midrule
FateZero~\cite{qi2023fatezero}                    & 2023                  & -                                               & CLIP                            & \color{red}{\ding{56}}                               & Diffusion                 & Training-free learning                                                                        & \color{green}{\ding{52}}            & \color{red}{\ding{56}}            & University of Science and Technology of China      \\ \midrule
CogVideoX~\cite{yang2024cogvideox}                   & 2024                  & 2B/5B & 3D Causal VAE                   & \color{red}{\ding{56}}                               & DiT                       & Image and video joint learning                                                                & \color{green}{\ding{52}}            & \color{green}{\ding{52}}            & Tsinghua University                                                                           \\ \midrule
HunyuanVideo~\cite{kong2024hunyuanvideo}                & 2024                  & 13B                                             & Causal VAE                      & \color{red}{\ding{56}}                               & DiT                       & Progressive learning                                                                          & \color{green}{\ding{52}}            & \color{green}{\ding{52}}            & Tencent                                                                                       \\ \midrule
MotionAura~\cite{susladkar2024motionaura}                  & 2024                  & 0.89B/1.94B/3.12B                               & 3D-MBQ-VAE                      & \color{green}{\ding{52}}                               & DiT                       & Image and video joint learning & \color{green}{\ding{52}}            & \color{red}{\ding{56}}            & Northwestern University                                                                       \\ \midrule
LARP~\cite{wang2024larp}                        & 2024                  & 805M                                            & Learnable token                 & \color{green}{\ding{52}}                               & Autoregressive            & Direct learning                                                                               & \color{green}{\ding{52}}            & \color{red}{\ding{56}}            & University of Maryland                                                                        \\ \midrule
xGen-VideoSyn-1~\cite{qin2024xgen}             & 2024                  & 731M                                            & 3D VAE                          & \color{green}{\ding{52}}                               & DiT                       & Direct learning                                                                               & \color{green}{\ding{52}}            & \color{red}{\ding{56}}            & Salesforce AI Research                                                                        \\ \midrule
DiCoDe~\cite{li2024dicode}                      & 2024                  & 3B                                              & Deep token                      & \color{green}{\ding{52}}                               & Autoregressive            & Progressive learning                                                                          & \color{red}{\ding{56}}            & \color{green}{\ding{52}}            & University of Hong Kong                                                                       \\ \midrule
OmniTokenizer~\cite{wang2024omnitokenizer}               & 2024                  & 227M/650M                                       & Feature decoupling              & \color{green}{\ding{52}}                               & Diffusion                 & Progressive learning                                                                          & \color{red}{\ding{56}}            & \color{green}{\ding{52}}            & Fudan University                                                                              \\ \midrule
SweetTok~\cite{tan2024sweettokenizer}                    & 2024                  & 2B                                              & DQAE                            & \color{green}{\ding{52}}                               & Autoregressive            & Transfer learning                                                                             & \color{green}{\ding{52}}            & \color{green}{\ding{52}}            & Kuaishou Technology                                                                           \\ \midrule
LTX-Video~\cite{hacohen2024ltx}                   & 2024                  & 2B                                              & 3D-VAE                          & \color{red}{\ding{56}}                               & DiT                       & Image and video joint learning                                                                & \color{green}{\ding{52}}            & \color{green}{\ding{52}}            & Lightricks                                                                                    \\ \midrule
FreeLong~\cite{lu2024freelong}                         & 2024                  & -                                               & -                             & \color{red}{\ding{56}}                               & Diffusion                 & Training-free learning                                                                          & \color{green}{\ding{52}}            & \color{red}{\ding{56}}            & University of Technology Sydney        \\ \midrule
Lumiere~\cite{bar2024lumiere}                     & 2024                  & -                                               & ST-Unet                         & \color{red}{\ding{56}}                               & Diffusion                 & -                                                                                             & \color{green}{\ding{52}}            & \color{green}{\ding{52}}            & Google Research                                                                               \\ \midrule
CMD~\cite{yu2024efficient}                         & 2024                  & 1.6B                                            & TimeSFormer                     & \color{red}{\ding{56}}                               & Diffusion                 & Direct learning                                                                               & \color{green}{\ding{52}}            & \color{red}{\ding{56}}            & KAIST                                                                                         \\ \midrule
TF-T2V~\cite{wang2024recipe}                      & 2024                  & 1.8B                                            & -                               & \color{red}{\ding{56}}                               & Diffusion                 & Image and video joint learning                                                                & \color{green}{\ding{52}}            & \color{red}{\ding{56}}            & Huazhong University of Science and Technology     \\ \midrule
HiGen~\cite{qing2024hierarchical}                       & 2024                  & -                                               & VAE                             & \color{red}{\ding{56}}                               & Diffusion                 & Image and video joint learning                                                                & \color{green}{\ding{52}}            & \color{red}{\ding{56}}            & Huazhong University of Science and Technology     \\ \midrule
Loong~\cite{wang2024loong}                       & 2024                  & 0.7B/3B/7B                                      & VQGAN                           & \color{green}{\ding{52}}                               & Autoregressive            & Progressive learning                                                                          & \color{green}{\ding{52}}            & \color{red}{\ding{56}}            & University of Hong Kong                                                                       \\ \midrule
Emu3~\cite{wang2024emu3}                        & 2024                  & 8B                                              & VQGAN                           & \color{green}{\ding{52}}                               & Autoregressive            & Image and video joint learning                                                                & \color{green}{\ding{52}}            & \color{red}{\ding{56}}            & BAAI                                                                                          \\ \midrule
Diffusion Forcing~\cite{chen2024diffusion}           & 2024                  & -                                               & VAE                             & \color{red}{\ding{56}}                               & Diffusion                 & Direct learning                                                                               & \color{green}{\ding{52}}            & \color{red}{\ding{56}}            & MIT                                                                                           \\ \midrule
ART-V~\cite{weng2024art}                       & 2024                  & -                                               & -                               & \color{red}{\ding{56}}                               & Autoregressive            & Progressive learning                                                                          & \color{green}{\ding{52}}            & \color{red}{\ding{56}}            & University of Science and Technology of China      \\ \midrule
NOVA~\cite{deng2024autoregressive}                        & 2024                  & 0.3B/0.6B/1.4B                                  & 3D VAE                          & \color{red}{\ding{56}}                               & Autoregressive            & Transfer learning                                                                             & \color{green}{\ding{52}}            & \color{red}{\ding{56}}            & Beijing University of Posts and Telecommunications \\ \midrule
MAGVIT~\cite{yu2023magvit}                      & 2024                  & 128M                                            & 3D VQ-VAE                       & \color{green}{\ding{52}}                               & Autoregressive            & Direct learning                                                                               & \color{green}{\ding{52}}            & \color{green}{\ding{52}}            & Carnegie Mellon University                                                                    \\ \midrule
MarDini~\cite{liu2024mardini}                     & 2024                  & 1.6B/3.4B                                       & VAE                             & \color{green}{\ding{52}}                               & Autoregressive            & Progressive learning                                                                          & \color{green}{\ding{52}}            & \color{green}{\ding{52}}            & Meta AI                                                                                       \\ \midrule
Emu Video~\cite{girdhar2024emuvideofactorizingtexttovideo}                   & 2024                  & 4.4B                                            & T5, CLIP, VAE                   & \color{red}{\ding{56}}                               & Diffusion                 & Progressive learning                                                                          & \color{green}{\ding{52}}            & \color{green}{\ding{52}}            & GenAI                                                                                         \\ \midrule
Factorized-Dreamer~\cite{yang2024factorizeddreamertraininghighqualityvideo}          & 2024                  & -                                               & T5, CLIP, VAE                   & \color{red}{\ding{56}}                               & Diffusion                 & Progressive learning                                                                          & \color{green}{\ding{52}}            & \color{green}{\ding{52}}            & ByteDance Inc                                                                                 \\ \midrule
DynamiCrafter~\cite{xing2024dynamicrafter}               & 2024                  & -                                               & CLIP, VAE                       & \color{red}{\ding{56}}                               & Diffusion                 & Transfer learning                                                                             & \color{green}{\ding{52}}            & \color{green}{\ding{52}}            & Chinese University of Hong Kong                  \\ \midrule
Framer~\cite{wang2024framer}                      & 2024                  & -                                               & VAE                             & \color{red}{\ding{56}}                               & Diffusion                 & Transfer learning                                                                             & \color{red}{\ding{56}}            & \color{green}{\ding{52}}            & Zhejiang University                                                                           \\ \midrule
FreeNoise~\cite{qiufreenoise}                        & 2024                  & -                                               & -                             & \color{red}{\ding{56}}                               & Diffusion                 & Training-free learning                                                                       & \color{green}{\ding{52}}            & \color{red}{\ding{56}}            & Nanyang Technological University              \\ \midrule
Genie~\cite{bruce2024genie}                       & 2024                  & 11B                                             & VQ-VAE                          & \color{green}{\ding{52}}                               & Autoregressive            & Reward feedback learning                                                                      & \color{green}{\ding{52}}            & \color{green}{\ding{52}}            & Google DeepMind                                                                               \\ \midrule
GameGen-X~\cite{che2024gamegen}                   & 2024                  & -                                               & 3D VAE, T5                      & \color{red}{\ding{56}}                               & DiT                       & Local module adaptation                                                                       & \color{green}{\ding{52}}            & \color{green}{\ding{52}}            & Hong Kong University of Science and Technology    \\ \midrule
StyleMaster~\cite{ye2024stylemaster}                 & 2024                  & -                                               & 3D VAE, CLIP                    & \color{red}{\ding{56}}                               & DiT                       & Local module adaptation                                                                       & \color{red}{\ding{56}}            & \color{green}{\ding{52}}            & Hong Kong University of Science and Technology     \\ \midrule
Show-1~\cite{zhang2025show1marryingpixellatent}                      & 2024                  & 6B                                              & CLIP, T5                        & \color{red}{\ding{56}}                               & Diffusion                 & Transfer learning                                                                             & \color{green}{\ding{52}}            & \color{red}{\ding{56}}            & National University of Singapore                   \\ \midrule
Sora~\cite{openai2024sora}                        & 2024                  & -                                               & VQ-VAE                          & \color{green}{\ding{52}}                               & DiT                       & -                                                                                             & \color{green}{\ding{52}}            & \color{red}{\ding{56}}            & OpenAI                                                                                        \\ \midrule
Ruyi~\cite{createai2024ruyi}                        & 2024                  & 7B                                              & Causal VAE, CLIP                & \color{green}{\ding{52}}                               & DiT                       & Transfer learning                                                                             & \color{green}{\ding{52}}            & \color{green}{\ding{52}}            & TuSimple                                                                                      \\ \midrule
InstructVideo~\cite{yuan2024instructvideo}               & 2024                  & -                                               & VQGAN                           & \color{green}{\ding{52}}                               & Diffusion                 & Reward feedback learning                                                                      & \color{green}{\ding{52}}            & \color{red}{\ding{56}}            & Zhejiang University                                                                           \\ \midrule
SimDA~\cite{xing2024simda}                       & 2024                  & 1.08B                                           & VAE                             & \color{red}{\ding{56}}                               & Diffusion                 & Local module adaptation                                                                       & \color{green}{\ding{52}}            & \color{red}{\ding{56}}            & Fudan University                                                                              \\ \midrule
Movie Gen~\cite{polyak2024movie}                   & 2024                  & 30B                                             & TAE                             & \color{red}{\ding{56}}                               & DiT                       & Image and video joint learning                                                                & \color{green}{\ding{52}}            & \color{red}{\ding{56}}            & Meta AI                                                                                       \\ \midrule
Wan~\cite{wan2025wan}                         & 2025                  & 1.3B/14B                                        & 3D Causal VAE                   & \color{green}{\ding{52}}                               & DiT                       & Image and video joint learning                                                                & \color{green}{\ding{52}}            & \color{green}{\ding{52}}            & Alibaba                                                                                       \\ \midrule
VidTwin~\cite{wang2025vidtwin}                     & 2025                  & -                                               & Feature decoupling              & \color{red}{\ding{56}}                               & Diffusion                 & Direct learning                                                                               & \color{green}{\ding{52}}            & \color{green}{\ding{52}}            & Peking University                                                                             \\ \midrule
Phys-AR~\cite{lin2025reasoning}                     & 2025                  & -                                               & DDT token                       & \color{green}{\ding{52}}                               & Autoregressive            & Reward feedback learning                                                                      & \color{red}{\ding{56}}            & \color{green}{\ding{52}}            & Zhejiang University                                                                           \\ \midrule
MAGI-1~\cite{ai2025magi1autoregressivevideogeneration}                      & 2025                  & 4.5B/24B                                        & VAE                             & \color{green}{\ding{52}}                               & Autoregressive            & Model distillation                                                                            & \color{green}{\ding{52}}            & \color{green}{\ding{52}}            & Sand AI                                                                                       \\ \midrule
MinT~\cite{wu2025mind}                        & 2025                  & -                                               & -                               & \color{red}{\ding{56}}                               & DiT                       & Transfer learning                                                                             & \color{green}{\ding{52}}            & \color{red}{\ding{56}}            & Snap Research                                                                                 \\ \midrule
VideoWorld~\cite{ren2025videoworldexploringknowledgelearning}                  & 2025                  & 0.3B                                            & Causal VQ-VAE                   & \color{green}{\ding{52}}                               & Autoregressive            & Local module adaptation                                                                       & \color{green}{\ding{52}}            & \color{green}{\ding{52}}            & Beijing Jiaotong University                                                                   \\ \midrule
Goku~\cite{chen2025gokuflowbasedvideo}                        & 2025                  & 1B/2B/8B                                        & 3D VAE                          & \color{red}{\ding{56}}                               & DiT                       & Image and video joint learning                                                                & \color{green}{\ding{52}}            & \color{green}{\ding{52}}            & The University of Hong Kong                                                                   \\ \midrule
EasyControl~\cite{wang2025easycontrol}                 & 2025                  & -                                               & VAE                             & \color{red}{\ding{56}}                               & DiT                       & Local module adaptation                                                                       & \color{green}{\ding{52}}            & \color{green}{\ding{52}}            & Sun Yat-sen University                                                                        \\ \midrule
VideoMerge~\cite{zhang2025videomerge}                  & 2025                  & -                                               & T5, CLIP                        & \color{red}{\ding{56}}                               & Diffusion                 & Training-free learning                                                                        & \color{green}{\ding{52}}            & \color{red}{\ding{56}}            & University of Central Florida                                                                 \\ \midrule
MOVi~\cite{rahman2025movi}                        & 2025                  & -                                               & -                               & \color{red}{\ding{56}}                               & Diffusion                 & Training-free learning                                                                        & \color{green}{\ding{52}}            & \color{red}{\ding{56}}            & Johns Hopkins University                                                                      \\ \midrule
Magic 1-For-1~\cite{yi2025magic}               & 2025                  & 13B                                             & CLIP, LLaVA-LLaMA               & \color{green}{\ding{52}}                               & Diffusion                 & Model distillation                                                                            & \color{green}{\ding{52}}            & \color{green}{\ding{52}}            & Peking University                                                                             \\ \midrule
MagicDistillation~\cite{shao2025magicdistillation}           & 2025                  & -                                               & VAE                             & \color{red}{\ding{56}}                               & DiT                       & Model distillation                                                                            & \color{red}{\ding{56}}            & \color{green}{\ding{52}}            & Hong Kong University of Science and Technology     \\ \midrule
LongDiff~\cite{li2025longdiff}           & 2025                  & -                                               & -                             & \color{red}{\ding{56}}                               & Diffusion                       & Training-free learning                                                                            & \color{green}{\ding{52}}            & \color{red}{\ding{56}}            & Lancaster University     \\ \bottomrule
\end{tabular}
}
\caption{Summary of Video Works}
\label{tab:Summary of Video Works}
\end{table*}

\section{Cases of Spatiotemporal Inconsistency}
\begin{figure*}[t!]
        
            \centering
             \scalebox{1}{
            \includegraphics[width=\textwidth]{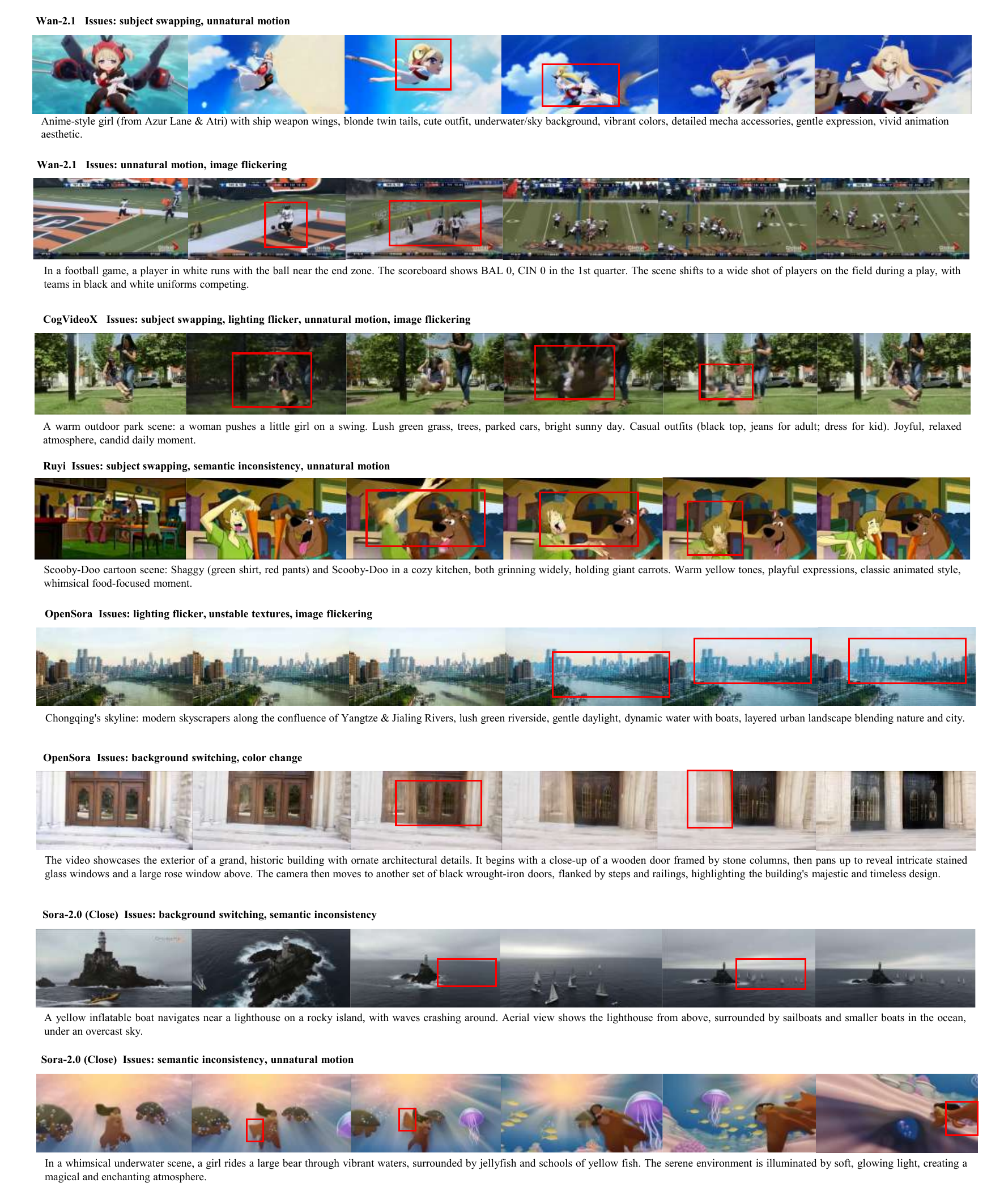}
            }
            
            \caption{Cases of Spatiotemporal Inconsistency.}
            \label{Fig:Cases of Spatiotemporal Inconsistency} 
            \vspace{-0.1cm}
        \end{figure*}

We selected four open-source models (Wan-2.1~\cite{wan2025wan}, CogVideoX~\cite{yang2024cogvideox}, Ruyi~\cite{createai2024ruyi}, OpenSora~\cite{peng2025open, zheng2024open}) and one closed-source model (Sora-2.0~\cite{sora}) for spatiotemporal consistency testing. The results are shown in Figure \ref{Fig:Cases of Spatiotemporal Inconsistency}, with identified issues annotated within the cases.

\section{Applications for Spatiotemporal Consistent Video Generation}
\begin{figure*}[t!]
        
            \centering
             \scalebox{1}{
            \includegraphics[width=\textwidth]{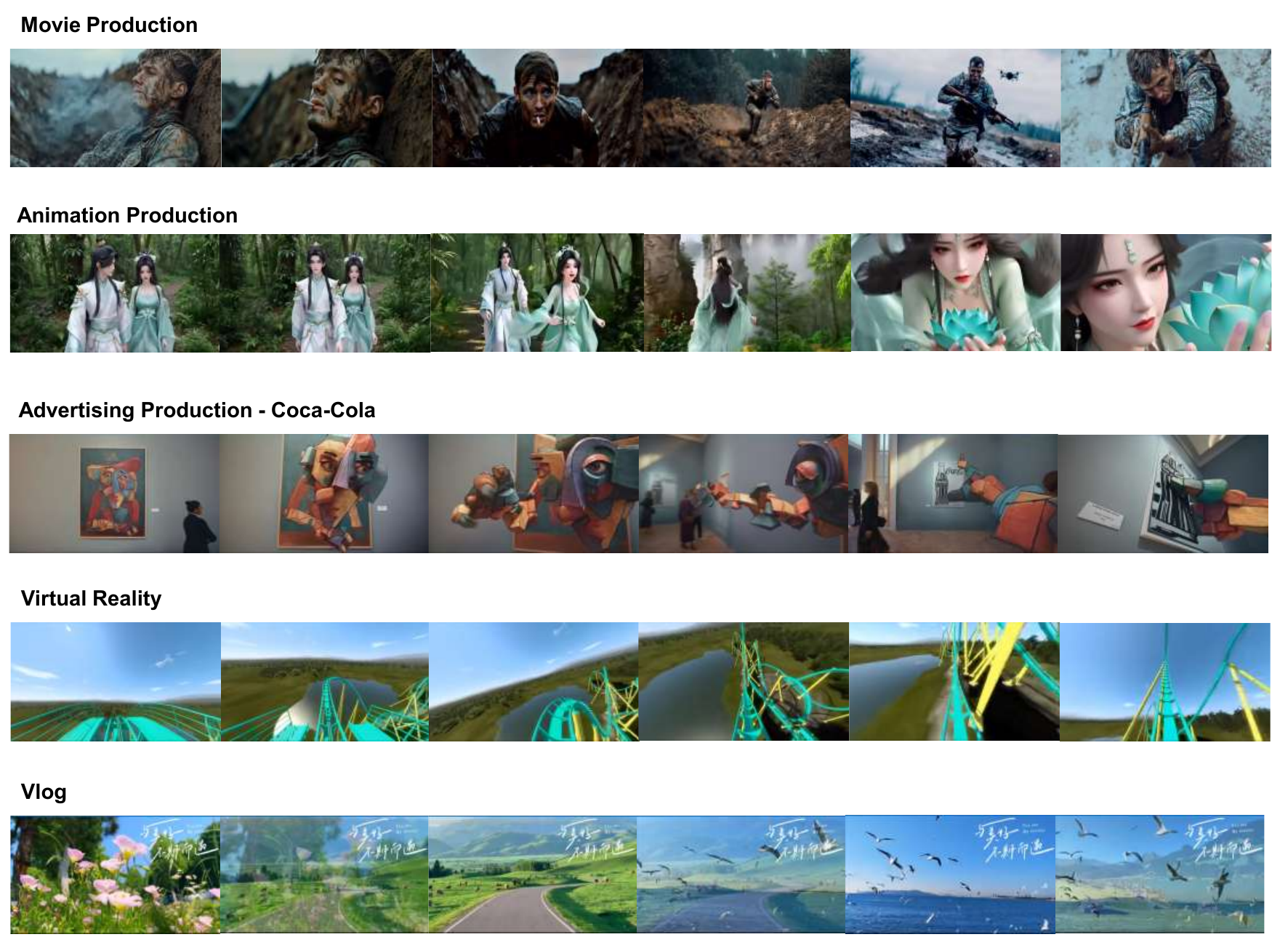}
            }
           
            \caption{Applications for Video Generation.}
            \label{Fig:Applications for Video Generation} 
            \vspace{-0.1cm}
        \end{figure*}

Spatiotemporal consistency has evolved from a academic research topic into a pivotal productivity tool driving transformation across multiple visual content industries. It not only resolves efficiency bottlenecks and cost issues in traditional production methods but also pioneers novel forms of creative expression. It plays a crucial role in fields such as movie production, animation production, advertising creation, virtual reality, and Vlog making. Some examples are shown in Figure \ref{Fig:Applications for Video Generation}.

In movie production, spatiotemporal consistency ensures seamless integration of characters with environments in dynamic scenes. Even within complex tracking shots or special effects composites, subjects maintain stable spatial relationships and temporal continuity with backgrounds, substantially enhancing visual realism.
Within animation production, this technology generates smoother motion sequences with physically accurate character movement, significantly reducing the workload associated with traditional hand-drawn or keyframe animation. For advertising production, spatiotemporal consistency maintains product form stability and natural lighting within dynamic scenes. Even during camera angle shifts or scene transitions, product imagery remains prominently and authentically presented, effectively conveying brand messaging. In virtual reality, this technology guarantees continuous, stable environments during user exploration. Perspective changes occur without abrupt jumps or distortion, which is paramount for immersion. For Vlog making, spatiotemporal consistency empowers creators to effortlessly implement background replacement, dynamic texturing, and style transfer. This enables ordinary users to rapidly generate professionally consistent video content, significantly lowering the barrier to personal creative expression.

In the future, as video generation continues to advance, spatiotemporal consistency technology is poised to evolve from ‘passively maintaining consistency’ to ‘actively constructing consistency’. This will enable not only the generation of coherent content but also the proactive planning of plausible spatiotemporal progression based on semantic and physical rules. Such capabilities promise to unlock greater potential in fields such as simulation training and medical visualization. This technology is increasingly becoming the pivotal bridge connecting virtual content generation with the rules of the real world, driving an ongoing intelligent transformation in the paradigm of visual content creation.

\end{document}